%% file: main.tex
\documentclass[review,12pt]{elsarticle}

\usepackage[utf8]{inputenc}
\usepackage[T1]{fontenc}
\usepackage{amsmath,amssymb,amsfonts}
\usepackage{mathtools}
\usepackage{bm}
\usepackage{graphicx}
\usepackage{xcolor}
\usepackage{booktabs}
\usepackage{tabularx}
\usepackage{rotating}
\usepackage{multirow}
\usepackage{array}
\usepackage{makecell}
\usepackage{hyperref}
\usepackage{cleveref}
\usepackage{algorithm}
\usepackage{algorithmic}
\usepackage{url}
\usepackage{lineno}
\usepackage{subcaption}
\usepackage{float}
\usepackage{float}           % برای [H]
\usepackage{algorithm}
\usepackage{algorithmic}
\usepackage{subcaption}
\usepackage{orcidlink}
\usepackage{adjustbox}
\usepackage{rotating}
\hypersetup{
  colorlinks=true,
  linkcolor=black,
  citecolor=blue,
  urlcolor=blue,
  pdftitle={Drain-Vortex Optimization},
  pdfauthor={Author Names}
}

\journal{Expert Systems with Applications}

\begin{document}

% =====================================================================
%                              FRONTMATTER
% =====================================================================
\begin{frontmatter}

\title{Drain-Vortex Optimization: A Population-Based Metaheuristic Inspired by Multi-Drain Free-Vortex Flow}

% Authors, affiliations, and corresponding author go here.
% Replace with the actual author list before submission.
\author[ad1]{Mohsen Omidi\,\orcidlink{0000-0002-0691-4183}}
\ead{Mohsen.Omidi@TUDublin.ie}

\author[ad1]{Brian Vaughan\,\orcidlink{0000-0002-5825-8990}}

%\author[ad1]{Corresponding Author Name\corref{cor1}}
%\ead{Brian.Vaughan@TUDublin.ie}

\cortext[cor1]{Corresponding author.}

\affiliation[ad1]{
  organization={Technological University Dublin},
  addressline={Grangegorman Lower},
  city={Dublin},
  postcode={D07 C972},
  country={Ireland}
}

% ----------- Abstract and keywords -----------
\input{abstract.tex}

\end{frontmatter}

% =====================================================================
%                                BODY
% =====================================================================

\input{sections/section_1_introduction.tex}

\input{sections/section_2_related_work.tex}

\input{sections/section_3_inspiration.tex}

\input{sections/section_4_algorithm.tex}

\input{sections/section_5_experimental_setup.tex}

% Tables that are referenced from the experimental setup section.
% (Most journals prefer floats placed near the first reference.)
\input{tables/table_experimental_protocol.tex}

\input{tables/table_algorithm_settings.tex}
\input{tables/table_reporting_convention.tex}

\input{sections/section_6_results.tex}

% Tables for results.
\input{tables/table_cec2022_results.tex}
\input{tables/table_cec2017_D30_results.tex}
\input{tables/table_cec2017_D50_results.tex}
\input{tables/table_classical_scalable_D30_results.tex}
\input{tables/table_classical_scalable_D100_results.tex}
\input{tables/table_classical_fixed_dimensional_results.tex}
\input{tables/table_engineering_best_feasible_results.tex}
\input{tables/table_convergence_final_results.tex}
\input{tables/table_ablation_cec2017_subset_results.tex}
\input{tables/table_friedman.tex}
\input{tables/table_wilcoxon.tex}
\clearpage

\input{sections/section_7_conclusion.tex}

% =====================================================================
%                          DECLARATIONS (optional)
% =====================================================================

\section*{Declaration of competing interest}
The authors declare that they have no known competing financial interests or personal relationships that could have appeared to influence the work reported in this paper.

\section*{Code availability}

The source code, experimental scripts, processed results, and generated LaTeX tables are publicly available at:
\url{https://github.com/Mohsen-Omidi/drain-vortex-optimization}.

% =====================================================================
%                            BIBLIOGRAPHY
% =====================================================================
\bibliographystyle{elsarticle-num}
\bibliography{refs}

\end{document}

%% file: abstract.tex
\begin{abstract}
This paper proposes Drain-Vortex Optimization (DVO), a population-based
metaheuristic for continuous optimization. DVO models each candidate
solution as a particle moving in a multi-drain vortex field. Its update
rule decomposes motion into radial attraction toward selected drain
centres and tangential rotation governed by a regularized free-vortex
law. A three-phase mechanism switches between far-field exploration,
spiral inward motion, and localized core exploitation according to the
normalized distance to the assigned drain. The method also uses
adaptive spiral exploitation, population-level vortex basin assignment,
and optional stochastic basin switching to support structured diversity.

DVO is evaluated against PSO, GWO, WOA, SCA, AOA, EO, and SVOA using
a calibration--validation protocol. CEC 2022 is used only to select
the final DVO configuration, while CEC 2017, classical functions, and
five constrained engineering design problems are used for out-of-sample
validation. On CEC 2017, DVO achieves the best mean $\log_{10}$ error
on 34 of 58 cases and the best Friedman average rank (1.66), and is
significantly better than every baseline under Holm-corrected Wilcoxon
tests. On CEC 2022, DVO obtains the best Friedman rank (2.13) and is
significantly better than five of the seven baselines; the differences
against PSO and SVOA are not significant. DVO is less competitive on
simple scalable classical functions and on small constrained
engineering designs, which clarifies its operating regime. The
algorithm is implemented in a vectorized GPU form that executes
independent runs in parallel.
\end{abstract}

%% file: sections/section_1_introduction.tex
\section{Introduction}
\label{sec:introduction}

Optimization problems are central to many fields of science and
engineering. From the design of mechanical components to the training
of machine learning models, from logistics planning to portfolio
management, the search for the best solution under a set of constraints
is a recurring challenge~\cite{boyd2004convex,bonnans2006numerical}.
When the objective function is differentiable, smooth, and
well-conditioned, classical gradient-based methods offer fast and
reliable convergence~\cite{bonnans2006numerical}. In practice, however,
many real-world problems do not satisfy these assumptions. The
objective function can be non-differentiable, multimodal, noisy, or
available only through expensive black-box
evaluations~\cite{audet2017derivative}. In such settings, gradient-based
methods are not directly applicable, and derivative-free optimization
techniques are preferred.

Among derivative-free methods, metaheuristic algorithms have become a
central tool in the optimization
literature~\cite{boussaid2013survey,hussain2019metaheuristic}. A
metaheuristic algorithm is a high-level procedure that uses random
sampling, population-based exploration, and a combination of local and
global search operators to identify high-quality solutions without
relying on derivative information. The most attractive property of
metaheuristics is their problem independence: a single algorithmic
template can be applied to a wide range of problems with minimal
customization. This generality is the reason why metaheuristics have
been adopted in many engineering domains, including structural
design~\cite{coello2002theoretical}, parameter
identification~\cite{khatibi2013pso}, feature
selection~\cite{xue2015survey}, and the tuning of complex control
systems~\cite{eberhart1995new,faramarzi2020equilibrium}.

The metaheuristic literature is large and continuously growing.
Classical algorithms include Genetic Algorithms
(GA)~\cite{holland1992adaptation}, Simulated Annealing
(SA)~\cite{kirkpatrick1983optimization}, and Particle Swarm Optimization
(PSO)~\cite{kennedy1995particle}. More recent algorithms have been
inspired by a wide variety of natural and physical phenomena. Examples
include the social hierarchy of grey wolves in the Grey Wolf Optimizer
(GWO)~\cite{mirjalili2014grey}, the bubble-net hunting strategy of
humpback whales in the Whale Optimization Algorithm
(WOA)~\cite{mirjalili2016whale}, the trigonometric oscillation in the
Sine Cosine Algorithm (SCA)~\cite{mirjalili2016sca}, the cooperative
behaviour of Harris hawks in the Harris Hawks Optimization
(HHO)~\cite{heidari2019harris}, the chemical equilibrium principle in
the Equilibrium Optimizer (EO)~\cite{faramarzi2020equilibrium}, the
basic operations of arithmetic in the Arithmetic Optimization Algorithm
(AOA)~\cite{abualigah2021arithmetic}, and the polar stratospheric flow
in the Stratospheric Vortex Optimization Algorithm
(SVOA)~\cite{he2026stratospheric}. Each of these algorithms
proposes a particular metaphor for the search process, and the
corresponding update rules introduce a specific balance between
exploration and exploitation.

Despite this rich literature, the design of metaheuristic algorithms
remains an active area of research, for two main reasons. First, the
No Free Lunch (NFL) theorem~\cite{wolpert2002no} states that no single
optimizer can outperform every other optimizer on every possible
objective function. Different problem classes therefore call for
different search strategies, and the discovery of an algorithm that
performs well on a particular class of problems does not preclude the
development of a more effective algorithm on a different class.
Second, even on widely used benchmark suites such as the IEEE CEC
competitions on bound-constrained numerical
optimization~\cite{awad2017cec2017,kumar2022cec2022}, the existing
metaheuristics still leave room for improvement, particularly on the
hybrid and composition functions that combine several landscape
components into a single objective. These functions are designed to
challenge the exploration--exploitation balance of an optimizer, and a
method that performs well on them is expected to generalize to
real-world problems with similarly heterogeneous landscapes.

A common limitation of many population-based metaheuristics lies in
the way the motion of an agent is constructed. In a typical update
rule, the new position of an agent is obtained by combining a single
attraction term, which pulls the agent towards a leading solution,
with a perturbation term that introduces
randomness~\cite{kennedy1995particle,mirjalili2014grey,mirjalili2016whale}.
The two terms are then mixed through coefficients that schedule the
transition between exploration and exploitation. This formulation has
produced many successful algorithms, but it conflates two distinct
aspects of the motion: the movement towards the leader (radial
direction) and the movement around the leader (tangential or
rotational direction). When these two aspects are mixed by a single
rule, it becomes difficult to control them independently, and the
algorithm can either converge prematurely (when the radial term
dominates) or fail to refine its solutions (when the perturbation
dominates). A search model that explicitly separates the radial and
tangential components, and that controls them with distinct rules
derived from a physically meaningful flow, has the potential to
improve this balance.

A second common limitation concerns the use of a single attractor or,
at most, of a small fixed number of leaders in the position update.
PSO uses a personal best and a global best per particle. GWO uses
three leaders ($\alpha$, $\beta$, $\delta$) but combines their
attractions into a single weighted average that effectively acts as a
single attractor. WOA uses a single leader during the encircling and
bubble-net phases. When the objective landscape contains many
promising basins of attraction, this single-attractor philosophy
biases the search towards a particular region and can cause the
population to lose track of competing basins. A multi-attractor
structure that maintains several promising regions simultaneously, and
that allows the population to redistribute itself across these regions
during the search, has been less explored in the metaheuristic
literature.

The present paper proposes Drain-Vortex Optimization (DVO), a new
metaheuristic algorithm that addresses both of the above limitations.
The algorithm is inspired by the rotational flow that develops when a
fluid drains from a basin through one or several outlets. This flow
pattern, observed in everyday phenomena such as the swirl that forms
above the outlet of a sink and the much larger whirl that develops
behind the spillway of a dam~\cite{lugt1983vortex,batchelor2000introduction},
exhibits three distinct flow regions and a continuous transition
between them. Far from the drain, the influence of the outlet is weak
and the motion is dominated by random surface disturbances. In an
intermediate region, the flow follows a spiral trajectory in which the
rotational component intensifies as the radial distance decreases.
Close to the drain, the flow is highly localized and rapidly evacuates
fluid through the outlet. These three regions provide a built-in
transition from exploration of the far-field region to exploitation of
the immediate neighbourhood of the drain, and they motivate the
three-phase motion model used in DVO. Furthermore, the multi-drain
extension of the same flow naturally supports a population that is
distributed across several basins of attraction, with a small
probabilistic mechanism for basin-to-basin information exchange.

Five contributions are presented in this paper.
\begin{itemize}
\item A drain-vortex-inspired motion model that decomposes the
trajectory of every agent into a radial component, derived from the
inward pressure gradient towards the drain, and a tangential
component, derived from the free-vortex law. The two components are
controlled separately and produce a spiral trajectory whose
rotational intensity grows as the agent approaches the drain.

\item An adaptive spiral exploitation schedule that maintains a
non-zero radial pressure throughout the search and prevents the
spiral component from degenerating into a pure rotation in the late
stages of the search. The schedule accelerates convergence in the
final stages without removing the rotational diversity of the spiral
phase.

\item A free-vortex swirl model based on a regularized form of the
free-vortex tangential velocity $v_\theta(r) = C/r$. The
regularization prevents numerical issues when an agent reaches the
immediate neighbourhood of a drain and preserves the physical scaling
of the swirl across dimensions.

\item A population-level vortex basin assignment in which every agent
is assigned to one of $K$ drains via a score that combines drain
quality and proximity. The assignment naturally distributes the
population across several basins of attraction without imposing a
hard partition.

\item An optional multi-vortex diversity mechanism that maintains
$K>1$ drain centres simultaneously and allows for stochastic
basin-to-basin transfers, modelled after the turbulent transfer
between vortex cells in a multi-drain basin. The contribution of
this optional component is examined empirically through an ablation
study, and it is presented as a diversity component whose value
depends on the problem class.
\end{itemize}

The proposed algorithm is evaluated on four benchmark suites that
together cover a broad range of optimization landscapes. The IEEE CEC
2022 single-objective bound-constrained suite is used as the
calibration benchmark for the parameter set of DVO. Three additional
suites serve as out-of-sample validation: the IEEE CEC 2017 suite
(twenty-nine functions in dimensions 30 and 50), a collection of
classical scalable and fixed-dimensional functions, and a collection
of five constrained engineering design problems. DVO is compared with
seven well-known baselines, namely
PSO~\cite{kennedy1995particle},
GWO~\cite{mirjalili2014grey},
WOA~\cite{mirjalili2016whale},
SCA~\cite{mirjalili2016sca},
AOA~\cite{abualigah2021arithmetic},
EO~\cite{faramarzi2020equilibrium}, and
SVOA~\cite{he2026stratospheric}, with reference parameter values
taken from the original papers or widely used implementations.

The experimental results show that DVO obtains the best mean
$\log_{10}$ error on 34 of the 58 cases of CEC 2017 and the best
Friedman average rank (1.66) on the same suite. The Wilcoxon
signed-rank test with Holm correction shows that DVO is significantly
better than every baseline on CEC 2017 at the 5\% level. On CEC 2022,
DVO obtains the best Friedman rank (2.13) and is significantly better
than five of the seven baselines; the differences against PSO and SVOA
are not significant. The Friedman test on CEC 2017 returns a $p$-value
of $4.36 \times 10^{-71}$, which provides strong evidence against
equivalent performance among the compared algorithms. On the classical
benchmark functions, DVO is mid-table because the structure of these
functions favours algorithms that contract aggressively around the
origin, and on the engineering subset DVO ranks fifth with a
feasibility rate of 98\%. The combined picture supports the position
that DVO is a promising optimizer for heterogeneous and
high-dimensional benchmark landscapes, while its relative weaknesses
on simpler classical and constrained engineering problems should be
considered when selecting it for a specific application. The honest
reporting of these relative weaknesses reflects the design philosophy
of the algorithm and provides the reader with a complete picture of
its operating regime.

The rest of the paper is organized as follows.
Section~\ref{sec:related_work} reviews the metaheuristic algorithms
that are most relevant to the proposed method.
Section~\ref{sec:inspiration} describes the inspiration of the
algorithm and presents the underlying physical model of drain-vortex
flow. Section~\ref{sec:algorithm} gives the mathematical formulation
of DVO, the pseudo-code, and the analysis of its computational
complexity. Section~\ref{sec:experimental_setup} describes the
experimental protocol and the benchmark suites.
Section~\ref{sec:results} reports the experimental results, the
convergence analysis, the ablation study, and the statistical
analysis. Section~\ref{sec:conclusion} concludes the paper and
outlines directions for future work.

%% file: sections/section_2_related_work.tex
% =====================================================================
% Section 2: Related Work
% =====================================================================

\section{Related Work}
\label{sec:related_work}

This section reviews the metaheuristic literature that is most relevant to the proposed algorithm. Section~\ref{subsec:rw_classification} introduces a brief classification of metaheuristics into evolution-based, swarm-based, and physics-based methods. Section~\ref{subsec:rw_swarm} reviews the swarm-based methods that are used as baselines in this study and describes their position update rules. Section~\ref{sec:related_physics} reviews the physics-based methods and identifies the few attempts that have used vortex or fluid-flow analogies for optimization. Section~\ref{subsec:rw_gap} closes the section by identifying the gap that the proposed algorithm fills.

\subsection{Classification of Metaheuristics}
\label{subsec:rw_classification}

Metaheuristic algorithms are commonly grouped into three large families based on the source of their inspiration~\cite{boussaid2013survey,hussain2019metaheuristic}. Evolution-based methods draw from biological evolution and reproduction. Genetic Algorithms (GA)~\cite{holland1992adaptation}, Differential Evolution (DE)~\cite{storn1997differential}, and Evolution Strategies (ES)~\cite{hansen2001completely} are the most prominent members of this family. Swarm-based methods draw from the collective behaviour of natural agents, including birds, insects, fish, and predators. Particle Swarm Optimization (PSO)~\cite{kennedy1995particle}, Ant Colony Optimization (ACO)~\cite{dorigo2006ant}, and the more recent Grey Wolf Optimizer (GWO)~\cite{mirjalili2014grey}, Whale Optimization Algorithm (WOA)~\cite{mirjalili2016whale}, and Harris Hawks Optimization (HHO)~\cite{heidari2019harris} fall in this category. Physics-based methods draw from physical phenomena and laws, including gravity, mechanics, electromagnetism, and chemistry. Simulated Annealing (SA)~\cite{kirkpatrick1983optimization}, Gravitational Search Algorithm (GSA)~\cite{rashedi2009gsa}, Henry Gas Solubility Optimization (HGSO)~\cite{hashim2019henry}, Equilibrium Optimizer (EO)~\cite{faramarzi2020equilibrium}, and Vortex Search (VS)~\cite{dogan2015new} are typical physics-based algorithms. A separate category is sometimes used for human-inspired methods such as Teaching-Learning-Based Optimization~\cite{rao2011teaching} and the Sine Cosine Algorithm (SCA)~\cite{mirjalili2016sca}, although the boundary between these categories is not always sharp.

The proposed algorithm belongs to the physics-based category, since it is inspired by the rotational flow that develops in a fluid that drains through one or several outlets. The motion model of the algorithm uses concepts from continuum fluid mechanics, including the free-vortex law and the radial--tangential decomposition of the velocity field~\cite{lugt1983vortex,batchelor2000introduction}. The next two subsections review the swarm-based methods and the physics-based methods most relevant to the proposed algorithm.

\subsection{Swarm-Based Methods}
\label{subsec:rw_swarm}

Particle Swarm Optimization is the canonical swarm-based metaheuristic~\cite{kennedy1995particle}. Each particle stores a position and a velocity, and the velocity is updated as a weighted combination of its current value, an attraction towards the personal best of the particle, and an attraction towards the global best of the swarm. The position is then obtained by integrating the velocity. PSO has produced many successful applications and has motivated a large number of follow-up studies~\cite{poli2007particle,zhan2009adaptive}. From the perspective of the present paper, the relevant feature of PSO is that the motion of every particle is the result of a single rule that mixes attraction and inertia, without an explicit decomposition into radial and tangential components.

The Grey Wolf Optimizer~\cite{mirjalili2014grey} models the social hierarchy of a wolf pack and the encircling, hunting, and attacking behaviours of grey wolves. Three leaders ($\alpha$, $\beta$, $\delta$) are maintained at every iteration, and the position of every other wolf is updated by averaging three attraction terms, one per leader. The convergence parameter $a$ decreases linearly from $2$ to $0$ over the course of the search and controls the balance between exploration and exploitation. Although GWO maintains three leaders, the position update collapses these leaders into a single weighted average, so that the agents effectively follow a single attractor. The Whale Optimization Algorithm~\cite{mirjalili2016whale} models the bubble-net hunting behaviour of humpback whales and uses two operators: an encircling operator that pulls the agent towards the best-found solution, and a spiral operator that follows a logarithmic curve around the best-found solution. The two operators are alternated by a probability of $0.5$, and the convergence parameter $a$ decreases linearly as in GWO. WOA introduces an explicit spiral motion in its update rule, but this spiral is two-dimensional and is parameterized by a single constant $b$ that controls the shape of the logarithmic curve. The spiral does not decompose the motion into a radial and a tangential component, and the spiral parameter is not updated during the search.

The Sine Cosine Algorithm~\cite{mirjalili2016sca} updates the position of every agent by adding a sinusoidal or cosinusoidal perturbation, with magnitudes scheduled by a coefficient that decreases linearly. SCA is computationally inexpensive and has been used as a baseline in many subsequent studies. The Arithmetic Optimization Algorithm~\cite{abualigah2021arithmetic} uses the four basic arithmetic operators to update the position of every agent. The exploration phase uses multiplication and division, and the exploitation phase uses addition and subtraction. AOA is a recent metaheuristic that has been validated on a wide range of benchmarks. The Harris Hawks Optimization~\cite{heidari2019harris} models the cooperative chase behaviour of Harris hawks and proposes several attack strategies that depend on the energy of the prey. HHO is one of the most cited swarm-based algorithms of the recent literature.

The common pattern across these swarm-based methods is the use of a single update rule that mixes attraction towards a leader (or a small group of leaders) with a perturbation term. The two components are not decomposed into a radial direction (towards the leader) and a tangential direction (around the leader), and the rotational component, when present, is restricted to a fixed two-dimensional curve as in WOA. The proposed DVO algorithm differs from these methods in that it explicitly separates the radial and tangential components of the motion, computes them with rules derived from a physically meaningful flow, and combines them through a closed-form spiral update.

\input{sections/section_2_3_physics_based.tex}

\subsection{Position of the Proposed Algorithm}
\label{subsec:rw_gap}

The review above identifies a gap in the metaheuristic literature.
To the best of our knowledge, none of the existing vortex-related
metaheuristics formulates the search process as a multi-drain drain-vortex
field with explicit radial attraction, regularized free-vortex tangential motion,
distance-dependent three-phase dynamics, and population-level drain-basin
assignment. Most swarm-based methods mix attraction and perturbation through
a single rule that does not decompose the motion. The few
physics-based methods that use fluid-flow analogies either operate on
a single solution and a contracting hypersphere, as in Vortex Search,
use a fixed two-dimensional spiral that is not derived from a
continuum fluid model, as in the swarm-based vortex variants, or rely
on an atmospheric polar-vortex model with Rossby-wave dynamics, as in
SVOA. None of these methods performs an explicit radial--tangential
decomposition of the agent motion or maintains several drain centres
simultaneously.

The proposed Drain-Vortex Optimization algorithm fills this gap by
formulating a population-based metaheuristic that is directly built
on the physics of multi-drain free-vortex flow. The motion of every
agent is decomposed into a radial pull and a tangential rotation.
The radial pull is scheduled by an adaptive spiral rule that
maintains a non-zero radial pressure throughout the search. The
tangential rotation is computed from a regularized form of the
free-vortex law and intensifies as the agent approaches the drain.
The population is distributed across $K$ drain centres through a
basin assignment that combines drain quality and proximity, and a
small probability of stochastic vortex switching allows agents to
occasionally move between drains. This combination of an explicit
radial--tangential decomposition, an adaptive spiral schedule, and a
multi-drain basin structure distinguishes DVO from SVOA and from the
other vortex-related methods reviewed above. The algorithm is
described in detail in the next section.

%% file: sections/section_2_3_physics_based.tex
\subsection{Physics-Based Methods}
\label{sec:related_physics}

The Gravitational Search Algorithm~\cite{rashedi2009gsa} models the
law of gravity and the laws of motion. Every agent has a mass that is
proportional to its fitness, and the population evolves under the
mutual gravitational attraction of all agents. GSA introduced the use
of physical interactions between agents in metaheuristic optimization.
The Equilibrium Optimizer~\cite{faramarzi2020equilibrium} models the
conservation of mass and the equilibrium state of a control volume in
a chemical reaction. Five equilibrium candidates are maintained at
every iteration, and every agent is updated towards a randomly
selected candidate. EO has obtained strong results on benchmark suites
and is used as one of the baselines in the present study. Henry Gas
Solubility Optimization~\cite{hashim2019henry} models the solubility
of gases in liquids according to Henry's law. The Atom Search
Optimization~\cite{zhao2019atom} models the interaction between atoms
based on the Lennard-Jones potential. These algorithms exploit
physical analogies that are different from the present work but
illustrate the productivity of the physics-based family.

A small number of metaheuristic algorithms have used vortex or
fluid-flow analogies. The Vortex Search algorithm~\cite{dogan2015new}
is a single-solution metaheuristic that samples points within a
contracting hypersphere centred at the current best solution. The
radius of the hypersphere is reduced by a deterministic schedule based
on an inverse incomplete gamma function, and the algorithm performs
the search by progressively narrowing the sampling region. Despite the
name, the Vortex Search algorithm does not model a fluid vortex.
There is no rotational motion in the update rule, no concept of
tangential velocity, and no analogue of the free-vortex law. The
vortex analogy is used metaphorically to describe a contracting
search region, and the resulting algorithm operates on a single
solution rather than a population. The
Vortex Swarm Optimization algorithm~\cite{elrahman2020vortex} couples a
swarm-based population with a vortex-shaped trajectory, but the
trajectory is a fixed two-dimensional spiral and the agent motion does
not derive from a continuum fluid model.

The Stratospheric Vortex Optimization Algorithm
(SVOA)~\cite{he2026stratospheric} models the polar vortex of the
stratosphere using a different physical model based on Rossby waves
and potential-vorticity gradients. SVOA is a population-based
metaheuristic that simulates the interaction of an air parcel with the
polar-vortex pressure field, and it has been reported to perform
strongly on modern competition suites. SVOA shares with DVO the
high-level idea of building a search algorithm on a multi-region
fluid-flow analogy, but the underlying physical model and the agent
update rules are very different. SVOA is driven by Rossby-wave
dynamics in a stratospheric flow, while DVO is driven by the radial
and tangential components of the drain-vortex flow with a regularized
free-vortex law for the swirl. SVOA does not perform an explicit
radial--tangential decomposition of the agent motion, and it does not
maintain several drain centres simultaneously. Because SVOA targets
the same class of optimization problems as DVO and has been shown to
be a strong competitor on shifted, rotated, and hybrid landscapes,
SVOA is included as one of the baselines of the experimental study.

Several vortex-related metaheuristics have been proposed, including the original Vortex Search algorithm and its modified, multi-objective, and population-based variants. More recent methods also include vortex-particle and atmospheric-vortex analogies. Table~\ref{tab:vortex_related_comparison} summarizes the main conceptual differences between these methods and the proposed DVO. The comparison in Table~\ref{tab:vortex_related_comparison} clarifies the
position of DVO relative to previous vortex-related methods. DVO is not
claimed to be the first vortex-inspired optimizer. Instead, its contribution
lies in the specific drain-vortex abstraction used to construct the search
dynamics. Previous vortex-related methods mainly rely on a contracting
search region, distributional sampling, fixed spiral trajectories, rotation
operators, or atmospheric-vortex analogies. In contrast, DVO models the
population as particles moving in a multi-drain basin, assigns each agent
to an active drain according to a quality--proximity score, and updates the
agent through an explicit radial--tangential decomposition governed by a
regularized free-vortex law. This combination of drain-basin assignment,
distance-dependent three-phase motion, and free-vortex tangential dynamics
is the main conceptual distinction of DVO.

\input{tables/table_vortex_related_comparison}

The comparison in Table~\ref{tab:vortex_related_comparison} clarifies that
DVO is not distinguished by the vortex metaphor alone. Its distinction lies
in the specific drain-vortex abstraction used to construct the search dynamics.
Previous vortex-related methods mainly rely on contracting sampling regions,
distributional search operators, fixed spiral trajectories, rotation-matrix
updates, or atmospheric-vortex dynamics. In contrast, DVO models the
population as particles moving in a multi-drain basin, assigns each agent to
an active drain according to a quality--proximity score, and updates the agent
through an explicit radial--tangential decomposition governed by a regularized
free-vortex law. To the best of our knowledge, this combination of drain-basin
assignment, distance-dependent three-phase motion, and free-vortex tangential
dynamics has not been used in previous vortex-related metaheuristics.

%% file: tables/table_vortex_related_comparison.tex
\begin{sidewaystable}[p]
\centering
\caption{Conceptual comparison between DVO and existing vortex-related metaheuristics.}
\label{tab:vortex_related_comparison}
\scriptsize
\setlength{\tabcolsep}{2.5pt}
\renewcommand{\arraystretch}{1.08}

\begin{adjustbox}{max width=\textheight, max totalheight=0.82\textwidth, keepaspectratio}
\begin{tabularx}{\textheight}{p{2.6cm} p{3.8cm} p{4.0cm} X}
\toprule
Method & Main physical analogy & Search structure / motion model & Main difference from DVO \\
\midrule

Vortex Search (VS)~\cite{dogan2015new}
& Vortical flow used as a metaphor for search contraction.
& Single-centre search that samples candidate solutions around a contracting region.
& VS does not model drain-vortex flow, does not use explicit radial--tangential motion, and does not maintain multiple active drain basins. \\

\midrule
Modified and enhanced VS variants~\cite{dougan2016modified,liu2023enhanced}
& Modified vortex-search sampling mechanisms.
& Usually retain the centre-based VS structure while changing distributions, candidate generation, or local enhancement.
& These methods improve the VS sampling mechanism, but they do not formulate the update through a regularized free-vortex tangential law or drain-basin assignment. \\

\midrule
Multi-objective VS~\cite{ozkis2017multiobjective}
& Vortex-search extension for Pareto optimization.
& Uses VS-style candidate generation with archive-based multi-objective selection.
& Its contribution is multi-objective selection rather than a drain-vortex physical update for single-objective continuous optimization. \\

\midrule
Population-based Vortex Search (PVS)~\cite{sag2022pvs}
& Population-based extension of VS.
& Uses a population, a location update operator, and polynomial mutation while retaining the VS radius-reduction mechanism.
& PVS is population-based, but it does not use drain-specific basin assignment, explicit radial attraction, free-vortex tangential rotation, or distance-dependent three-phase motion. \\

\midrule
Stratospheric Vortex Optimization Algorithm (SVOA)~\cite{he2026stratospheric}
& Polar stratospheric vortex dynamics.
& Combines potential-vorticity-gradient radial convergence, Rossby-wave-like motion, and sudden-stratospheric-warming perturbation.
& SVOA uses atmospheric-vortex dynamics, whereas DVO uses a drain-vortex basin model with explicit radial--tangential decomposition and stochastic basin switching. \\

\midrule
Proposed DVO
& Drain-vortex flow above basin outlets.
& Population-based multi-drain search with radial attraction, regularized free-vortex tangential rotation, and distance-dependent three-phase motion.
& DVO combines drain-basin assignment, explicit radial--tangential dynamics, adaptive spiral exploitation, and stochastic basin-to-basin switching in a single optimizer. \\

\bottomrule
\end{tabularx}
\end{adjustbox}

\end{sidewaystable}

%% file: sections/section_3_inspiration.tex
% =====================================================================
% Section 3: Inspiration -- Drain Vortex Physics
% =====================================================================

\section{Inspiration: Drain-Vortex Flow}
\label{sec:inspiration}

In this section, the inspiration of the proposed method is first discussed. Then, the relevant physical model is provided.

\subsection{Inspiration}
\label{subsec:inspiration_concept}

\begin{figure*}[!t]
  \centering
  \includegraphics[width=0.95\linewidth]{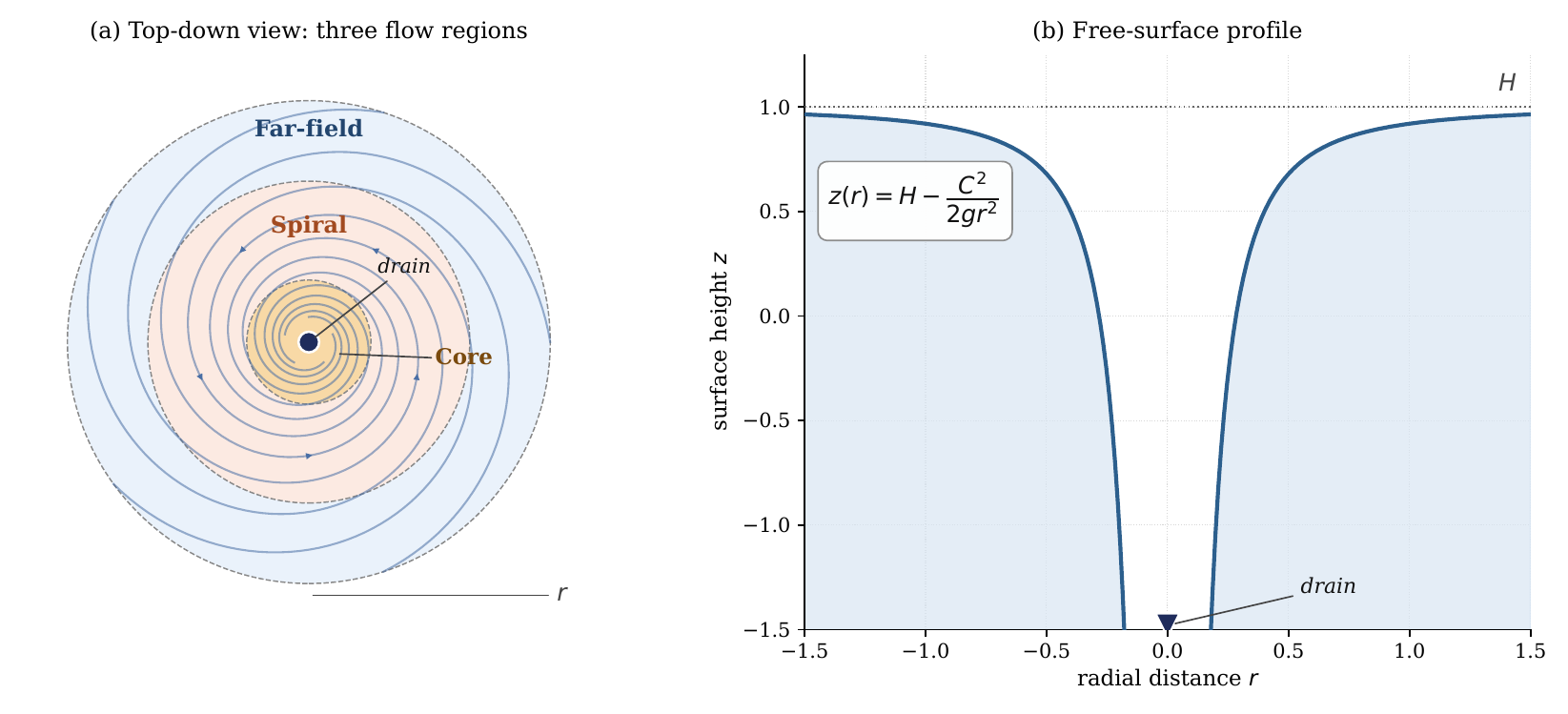}
  \caption{Inspiration of DVO from drain-vortex flow.
  (a) Top-down view of the three flow regions: a far-field where motion
  is mild, a spiral region where the rotational component intensifies as
  the radial distance decreases, and a core region of localized rapid
  evacuation. (b) Free-surface profile showing the hyperbolic depression
  predicted by Bernoulli's equation under the free-vortex assumption.}
  \label{fig:drain_vortex}
\end{figure*}

The drain vortex is a well-known phenomenon in fluid mechanics in which a fluid leaving a basin through a small outlet organizes itself into a rotating motion around the outlet \cite{batchelor2000introduction,lugt1983vortex}. The phenomenon is observed at very different scales, ranging from the small whirl that forms above the outlet of a sink, to the spiral motion above the spillway of a hydraulic dam, to large vortices observed in oceanographic and atmospheric flows. The persistence of the same flow pattern across these scales suggests that the underlying mechanism is general and is governed by a small number of physical parameters \cite{lugt1983vortex} as illustrated in Fig.~\ref{fig:drain_vortex}.

Of particular interest is the way in which a drain vortex naturally organizes the surrounding fluid into three regions with very different dynamic behaviour. The fluid far from the drain is only weakly affected by the outlet and moves slowly under the influence of the surrounding flow. The fluid in an intermediate annular region rotates around the drain and moves towards the outlet along a spiral path; this region is dominated by a strong tangential component of the velocity. The fluid very close to the drain rotates rapidly and is pulled into the outlet through a thin core region. The transition between these three regions is smooth and is governed by the radial distance from the drain centre \cite{lugt1983vortex,andersen2003anatomy}.

When several drains are present in the same basin, each drain produces its own local vortex and its own region of influence. A fluid parcel that lies close to a particular drain is mainly attracted by that drain, while a parcel located between two drains is influenced by both, in proportion to its distance from each. The flow field is therefore the superposition of several local vortex fields, and the basin is partitioned into regions of attraction whose boundaries are determined by the relative strengths and positions of the drains. It has also been observed that turbulent fluctuations in the bulk flow can occasionally transfer a fluid parcel from the basin of one drain to the basin of another \cite{andersen2003anatomy}. This transfer is rare but provides a mechanism for information to propagate between regions of the basin.

In the present work, the multi-drain behaviour and the three-region structure of a single drain vortex are abstracted into a mathematical search model for Drain-Vortex Optimization (DVO). The proposed mathematical model is presented in the following subsection.

\subsection{Mathematical model of the drain vortex}
\label{subsec:free_vortex}

The classical free-vortex model provides a compact mathematical description of the drain-vortex phenomenon \cite{batchelor2000introduction}. According to this model, the tangential component of the velocity at radial distance $r$ from the drain centre is given by
\begin{equation}
v_{\theta}(r) = \frac{C}{r},
\label{eq:free_vortex_speed}
\end{equation}
where $v_{\theta}$ is the tangential velocity, $r$ is the radial distance, and $C$ is a positive constant called the circulation strength. Equation~\eqref{eq:free_vortex_speed} captures the most important kinematic feature of a drain vortex: the closer a fluid element is to the drain centre, the higher its tangential velocity. As $r$ decreases, $v_{\theta}$ increases without bound in this idealized formulation. In a real fluid, viscous effects prevent this singularity and a thin core region of finite radius forms at the centre of the vortex \cite{lugt1983vortex}. The mathematical model used in DVO incorporates a regularized form of \eqref{eq:free_vortex_speed} in which a small core radius is added to the denominator in order to remove the singularity and improve numerical stability.

The free surface of the rotating fluid follows a related profile that can be obtained from the Bernoulli equation applied to the steady-state flow. If $H$ denotes the height of the fluid far from the drain and $g$ denotes the gravitational acceleration, the surface height at radial distance $r$ is given by \cite{andersen2003anatomy}
\begin{equation}
z(r) = H - \frac{C^{2}}{2 g r^{2}}.
\label{eq:free_vortex_surface}
\end{equation}
Equation~\eqref{eq:free_vortex_surface} predicts a hyperbolic depression of the free surface around the drain. The surface is almost flat for large values of $r$, develops a curved depression in the spiral region, and forms a deep narrow funnel in the vicinity of the drain. The depth of the funnel is controlled by the circulation strength $C$. A larger value of $C$ yields a deeper and wider funnel and therefore a more aggressive pull towards the drain centre.

Equations~\eqref{eq:free_vortex_speed} and~\eqref{eq:free_vortex_surface} together specify the kinematic structure of an idealized drain vortex. In the proposed DVO algorithm, \eqref{eq:free_vortex_speed} is used to compute the tangential component of the motion of a candidate solution as it spirals towards a vortex centre. Equation~\eqref{eq:free_vortex_surface} is used as a conceptual basis for the radial component of the motion, which becomes stronger as the candidate solution approaches the centre. To see the effects of these equations in an optimization context, two key observations may be noted.

First, the radial and tangential components of the motion can be decomposed and controlled independently. The radial component drives convergence towards a promising candidate solution and is therefore associated with exploitation. The tangential component preserves a rotational element in the trajectory and is therefore associated with diversity. The combination of the two components produces the spiral motion that characterizes a drain vortex.

Second, the magnitude of the tangential velocity in \eqref{eq:free_vortex_speed} grows monotonically as $r$ decreases. This property provides a built-in transition between exploration and exploitation as a candidate solution moves towards the vortex centre. Far from the vortex, the tangential motion is mild and the search is dominated by random disturbances. In the spiral region, the rotational component becomes significant and the search agent follows a structured curved trajectory. Near the vortex centre, the rotational component is strong and the search becomes highly localized. A single physical model therefore provides a smooth shift from global to local search without requiring an explicit external schedule.

\subsection{Multi-drain extension}
\label{subsec:multi_drain_extension}

\begin{figure}[!t]
  \centering
  \includegraphics[width=\linewidth]{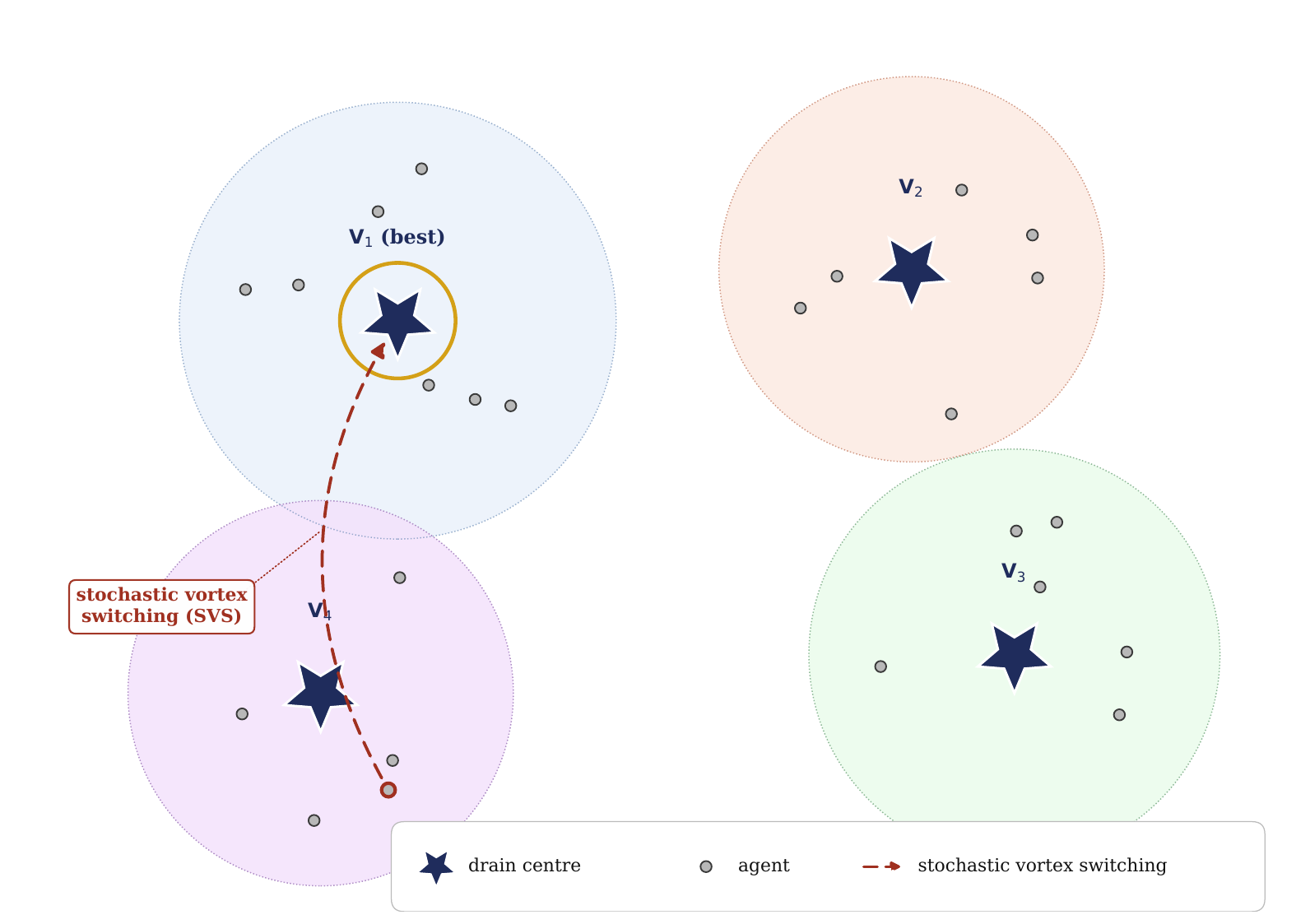}
  \caption{Multi-drain configuration. Each drain $\mathbf{V}_k$ produces
  its own basin of attraction (coloured halos). Agents are distributed
  across basins via a score that combines drain quality and proximity.
  The dashed arrow shows the stochastic vortex switching (SVS) mechanism,
  which transfers an agent from one basin to another with a small
  probability $p_{\mathrm{switch}}$.}
  \label{fig:multi_vortex}
\end{figure}

In a multi-drain configuration, each drain is treated as an independent local vortex, and the search agents are partitioned into groups according to the drain that exerts the strongest combined attraction. The attraction of a given drain on a given agent is determined by two factors: the quality of the drain, which is measured by the fitness of the corresponding candidate solution, and the distance between the drain and the agent. Drains that correspond to better candidate solutions exert a stronger pull, while drains that are closer to a given agent are more likely to capture it as illustrated in Fig.~\ref{fig:multi_vortex}.

In real flows, turbulent fluctuations can occasionally transfer a parcel from the basin of one drain to the basin of another \cite{andersen2003anatomy}. This effect is incorporated into DVO through a probabilistic switching mechanism that allows an agent to be reassigned, with a small probability, from its current drain to a different drain. The destination drain is sampled in proportion to the fitness-based attractiveness of the available drains. This mechanism prevents the population from collapsing into independent sub-populations around each drain and provides a controlled form of basin-to-basin information exchange.

The pseudo code and the detailed mathematical formulation of the DVO algorithm are presented in the next section. The algorithm uses one drain per leading candidate solution, computes the radial and tangential components of the motion from the regularized free-vortex model, and switches between three motion phases according to the radial distance of each agent.

%% file: sections/section_4_algorithm.tex
% =====================================================================
% Section 4: The DVO Algorithm
% =====================================================================

\section{The DVO Algorithm}
\label{sec:algorithm}

This section formulates the Drain-Vortex Optimization (DVO) algorithm. The flow phenomenon described in Section~\ref{sec:inspiration} is translated into a mathematical model for optimization in continuous search spaces. Section~\ref{subsec:overview} provides an overview of the algorithm. Section~\ref{subsec:formulation} presents the mathematical formulation of each component. The pseudo-code is given in Section~\ref{subsec:pseudocode}, and the computational complexity is analysed in Section~\ref{subsec:complexity}.

\subsection{Overview}
\label{subsec:overview}

In DVO, each candidate solution is treated as a water particle that moves through a basin containing $K$ drains. The drains are positioned at the $K$ best solutions found so far, and their positions are refreshed at every iteration through an elitist selection rule. At every iteration, each agent is assigned to one drain, and the motion of the agent is then determined by three components that together reproduce the drain-vortex flow described in the previous section. The first component is a far-field random walk with a weak drift towards the assigned drain. The second component is a spiral inward motion that combines a radial pull with a tangential rotation, with magnitudes derived from the free-vortex law. The third component is a localized core sampling around the drain. The component used for a given agent depends on its current normalized distance to the assigned drain. A small probability of stochastic vortex switching reassigns an agent from its current drain to a different drain, which models the turbulent transfer between vortex cells. Agents that remain stagnant in the core region for several iterations are re-launched into the search space through a L\'evy flight, which models the splash-out of a fluid parcel from the surface of the basin.

The mathematical model below uses the following notation. The search space is a $D$-dimensional bounded box $[\mathbf{lb}, \mathbf{ub}] \subset \mathbb{R}^{D}$. The population contains $N$ agents and the algorithm runs for a maximum of $T$ iterations. The position of agent $i$ at iteration $t$ is denoted by $\mathbf{X}_{i}^{t} \in \mathbb{R}^{D}$, and the position of drain $k$ is denoted by $\mathbf{V}_{k}^{t} \in \mathbb{R}^{D}$. The objective function $f$ is to be minimized. The Euclidean distance between agent $i$ and drain $k$ is $r_{i,k}^{t} = \|\mathbf{X}_{i}^{t} - \mathbf{V}_{k}^{t}\|$. The diameter of the search space, defined as $L = \|\mathbf{ub} - \mathbf{lb}\|$, is used to normalize this distance into $\rho_{i,k}^{t} = r_{i,k}^{t} / L$. The normalized distance $\rho_{i,k}^{t}$ takes values in $[0, 1]$ and removes the dependence of the geometry on the magnitude of the bounds.

\subsection{Mathematical Formulation}
\label{subsec:formulation}

\subsubsection{Initialization}
\label{subsubsec:init}

The initial population is sampled uniformly within the search space:
\begin{equation}
\mathbf{X}_{i}^{0} = \mathbf{lb} + \mathbf{r} \odot (\mathbf{ub} - \mathbf{lb}), \quad i = 1, \dots, N,
\label{eq:init}
\end{equation}
where $\mathbf{r} \in [0,1]^{D}$ is a vector of independent uniform random variables and $\odot$ denotes the element-wise (Hadamard) product. The objective function is then evaluated for every agent. The $K$ agents with the lowest fitness values become the initial drain positions $\mathbf{V}_{1}^{0}, \dots, \mathbf{V}_{K}^{0}$. The drains are stored in increasing order of fitness, so that $\mathbf{V}_{1}$ is always the best solution found so far and $\mathbf{V}_{K}$ is the worst of the $K$ retained drains.

\subsubsection{Drain Selection and Elitism}
\label{subsubsec:elitism}

At every iteration, the drain positions are recomputed from a candidate pool that combines the current population, the previous population, and the previous drain positions:
\begin{equation}
\mathcal{P}^{t} = \mathbf{X}^{t}\,\cup\,\mathbf{X}^{t-1}\,\cup\,\mathbf{V}^{t-1}.
\label{eq:elitism_pool}
\end{equation}
The $K$ best solutions of $\mathcal{P}^{t}$ are selected and their positions are assigned to $\mathbf{V}^{t}$. This elitist rule ensures that high-quality solutions discovered at any past iteration are preserved as drain positions, even if they are temporarily replaced by less promising candidates in the current population. The same rule also prevents the algorithm from losing track of the best-so-far region of the search space when a perturbation step deteriorates the population.

\subsubsection{Drain Probability and Assignment}
\label{subsubsec:assignment}

A vector of drain probabilities $\mathbf{P}^{t} = [P_{1}^{t}, \dots, P_{K}^{t}]$ is computed from the ranks of the drains. Drain $k$ is associated with rank $k-1$, where rank $0$ corresponds to the best drain. The probabilities are obtained from a softmax of the negated rank scores:
\begin{equation}
P_{k}^{t} = \dfrac{\exp\!\left(-\beta(t)\,(k-1)\,/\,(K-1)\right)}{\sum_{j=1}^{K} \exp\!\left(-\beta(t)\,(j-1)\,/\,(K-1)\right)}, \quad k = 1, \dots, K,
\label{eq:drain_probability}
\end{equation}
where $\beta(t)$ is a selection-pressure parameter that increases linearly from $\beta_{0}$ to $\beta_{1}$ over the course of the search:
\begin{equation}
\beta(t) = \beta_{0} + (\beta_{1} - \beta_{0})\,\frac{t}{T-1}.
\label{eq:beta_schedule}
\end{equation}
A small $\beta(t)$ produces a nearly uniform distribution that favours exploration across drains, whereas a large $\beta(t)$ concentrates the probability mass on the best drains and favours exploitation. This rank-based formulation is invariant under monotonic transformations of the objective function and remains numerically stable when objective values are large, very small, or possibly negative.

The assignment of agent $i$ to a drain is performed by maximizing a score that combines drain probability and proximity:
\begin{equation}
\mathrm{Score}_{i,k}^{t} = \dfrac{P_{k}^{t}}{\rho_{i,k}^{t} + \varepsilon},
\quad
k_{i}^{t} = \arg\max_{k}\,\mathrm{Score}_{i,k}^{t},
\label{eq:assignment_score}
\end{equation}
where $\varepsilon$ is a small positive constant that prevents division by zero. According to Eq.~\eqref{eq:assignment_score}, an agent is preferentially assigned to a high-quality drain that is also close to its current position. Agents that are equidistant from several drains are assigned to the drain with the highest probability, while agents that are very close to a single drain are assigned to that drain even if its probability is moderate. The assigned drain is denoted by $\mathbf{V}_{i}^{t} = \mathbf{V}_{k_{i}^{t}}^{t}$, the assigned distance by $r_{i}^{t} = r_{i,k_{i}^{t}}^{t}$, and the assigned normalized distance by $\rho_{i}^{t} = \rho_{i,k_{i}^{t}}^{t}$.

\subsubsection{Three Motion Phases}
\label{subsubsec:phases}

The motion of agent $i$ is determined by its current normalized distance $\rho_{i}^{t}$ to the assigned drain. Two thresholds $\rho_{\mathrm{far}}$ and $\rho_{\mathrm{near}}$, with $0 < \rho_{\mathrm{near}} < \rho_{\mathrm{far}} < 1$, partition the basin into three regions and select the corresponding motion phase:
\begin{equation}
\mathbf{X}_{i}^{t+1} =
\begin{cases}
\Phi_{\mathrm{far}}\!\left(\mathbf{X}_{i}^{t}, \mathbf{V}_{i}^{t}\right), & \rho_{i}^{t} > \rho_{\mathrm{far}},\\[2pt]
\Phi_{\mathrm{spiral}}\!\left(\mathbf{X}_{i}^{t}, \mathbf{V}_{i}^{t}, r_{i}^{t}, \rho_{i}^{t}\right), & \rho_{\mathrm{near}} < \rho_{i}^{t} \le \rho_{\mathrm{far}},\\[2pt]
\Phi_{\mathrm{core}}\!\left(\mathbf{V}_{i}^{t}\right), & \rho_{i}^{t} \le \rho_{\mathrm{near}}.
\end{cases}
\label{eq:phase_selection}
\end{equation}

\begin{figure}[!t]
  \centering
  \includegraphics[width=\linewidth]{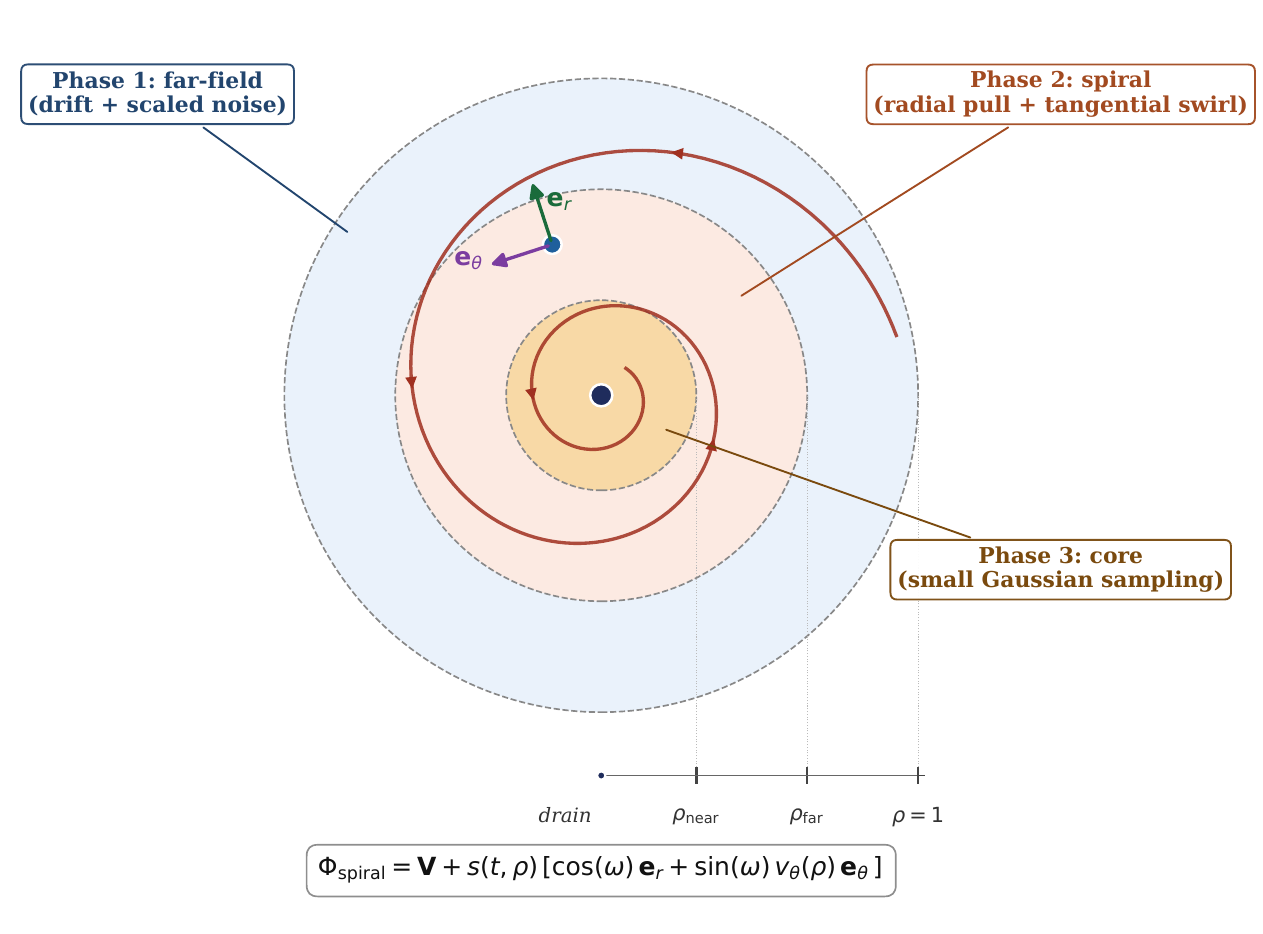}
  \caption{The three motion phases of an agent in DVO. In the far-field
  ($\rho > \rho_{\mathrm{far}}$), motion combines a weak drift toward
  the assigned drain with scaled noise. In the spiral region
  ($\rho_{\mathrm{near}} < \rho \le \rho_{\mathrm{far}}$), the agent
  follows a closed-form spiral combining a radial pull and a tangential
  rotation derived from the regularized free-vortex law. In the core
  ($\rho \le \rho_{\mathrm{near}}$), the agent samples positions from a
  small Gaussian cloud centred on the drain. The radial unit vector
  $\mathbf{e}_r$ and the tangential unit vector $\mathbf{e}_\theta$
  decompose the spiral motion into independent radial and rotational
  components.}
  \label{fig:three_phase}
\end{figure}
Fig.~\ref{fig:three_phase} shows the three motion regimes.
This piecewise rule reproduces the three flow regions of a drain vortex: far-field exploration, spiral inward motion, and core exploitation. The transition between phases is automatic and does not require a global iteration counter, since it depends only on the current geometry of the agent relative to its assigned drain.

\paragraph{Far-field phase.}
When the agent is far from its assigned drain, its motion is dominated by exploratory random walks combined with a weak drift towards the drain. The update rule is
\begin{equation}
\Phi_{\mathrm{far}}(\mathbf{X}_{i}^{t}, \mathbf{V}_{i}^{t}) = \mathbf{X}_{i}^{t} + \alpha_{1}\,\phi(t)\,(\mathbf{V}_{i}^{t} - \mathbf{X}_{i}^{t}) + \delta_{1}\,\dfrac{\phi(t)}{2}\,\dfrac{\mathbf{ub} - \mathbf{lb}}{\sqrt{D}}\,\boldsymbol{\eta}_{i}^{t},
\label{eq:phase1}
\end{equation}
where $\boldsymbol{\eta}_{i}^{t} \sim \mathcal{N}(\mathbf{0}, \mathbf{I}_{D})$ is a vector of independent standard normal samples, $\alpha_{1}$ is the drift coefficient, $\delta_{1}$ is the noise coefficient, and $\phi(t)$ is a control parameter that linearly decreases from $2$ to $0$:
\begin{equation}
\phi(t) = 2\,\left(1 - \frac{t}{T-1}\right).
\label{eq:phi_schedule}
\end{equation}
The noise term in Eq.~\eqref{eq:phase1} is rescaled by the variable range $(\mathbf{ub} - \mathbf{lb})$ and divided by $\sqrt{D}$, so that the typical magnitude of the random step is proportional to the size of the bounding box and is independent of the dimensionality of the search space. As iterations progress, $\phi(t)$ decreases and both the drift and the noise contributions vanish, which gradually narrows the exploration to the neighbourhood of the drain.

\paragraph{Spiral inward phase.}
When the agent is in the intermediate region, its motion follows a spiral trajectory that combines a radial pull and a tangential rotation. Let $\mathbf{e}_{r,i}^{t}$ be the unit vector pointing from the assigned drain to the agent and let $\mathbf{e}_{\theta,i}^{t}$ be a unit vector in $\mathbb{R}^{D}$ that is randomly drawn in the hyperplane perpendicular to $\mathbf{e}_{r,i}^{t}$:
\begin{equation}
\mathbf{e}_{r,i}^{t} = \dfrac{\mathbf{X}_{i}^{t} - \mathbf{V}_{i}^{t}}{r_{i}^{t} + \varepsilon},
\quad
\mathbf{e}_{\theta,i}^{t} \perp \mathbf{e}_{r,i}^{t},
\quad
\|\mathbf{e}_{\theta,i}^{t}\| = 1.
\label{eq:radial_tangential_basis}
\end{equation}
The tangential vector $\mathbf{e}_{\theta,i}^{t}$ is computed by drawing an auxiliary Gaussian vector, projecting it onto the hyperplane perpendicular to $\mathbf{e}_{r,i}^{t}$, and normalizing the result. The regularized free-vortex law is then used to obtain the magnitude of the tangential velocity:
\begin{equation}
v_{\theta}\!\left(\rho_{i}^{t}\right) = \min\!\left(\dfrac{C}{\rho_{i}^{t} + \rho_{\mathrm{core}}},\,v_{\max}\right),
\label{eq:vtheta}
\end{equation}
where $C$ is the circulation strength, $\rho_{\mathrm{core}}$ is a small positive constant that prevents the singularity at $\rho = 0$, and $v_{\max}$ is an upper bound that ensures numerical stability when an agent reaches the immediate neighbourhood of a drain. Equation~\eqref{eq:vtheta} is the regularized counterpart of the ideal free-vortex law of Eq.~\eqref{eq:free_vortex_speed} and is therefore directly inherited from the physical model of Section~\ref{sec:inspiration}.

The radial component of the spiral motion is obtained from a shrink factor that defines the new radial distance after one update step:
\begin{equation}
s(t, \rho_{i}^{t}) = \big(1 - \gamma\,c(t)\big)\,r_{i}^{t},
\quad
c(t) = c_{\min} + (1 - c_{\min})\,\dfrac{\phi(t)}{2}.
\label{eq:shrink_radial}
\end{equation}
The variable $c(t)$ schedules the radial pressure between an initial value of $1$ and a residual value $c_{\min}$ at the end of the search. This formulation is referred to as the \emph{adaptive spiral} schedule. It ensures that a non-zero radial pull is preserved at every iteration, which prevents the spiral component from degenerating into a pure rotation in the late stages of the search and accelerates convergence in the final stages of the search. The full update rule of the spiral phase is then
\begin{equation}
\Phi_{\mathrm{spiral}}\!\left(\mathbf{X}_{i}^{t}, \mathbf{V}_{i}^{t}, r_{i}^{t}, \rho_{i}^{t}\right) = \mathbf{V}_{i}^{t} + s(t, \rho_{i}^{t})\,\big[\cos(\omega_{i}^{t})\,\mathbf{e}_{r,i}^{t} + \sin(\omega_{i}^{t})\,v_{\theta}\!\left(\rho_{i}^{t}\right)\,\mathbf{e}_{\theta,i}^{t}\big],
\label{eq:phase2}
\end{equation}
where $\omega_{i}^{t} \sim \mathcal{U}(0, 2\pi)$ is a uniformly distributed phase angle. Equation~\eqref{eq:phase2} is the central component of DVO and combines several physical effects in a single closed-form rule. The radial pull, controlled by the shrink factor $s(t, \rho_{i}^{t})$, brings the agent closer to the drain at every iteration. The tangential rotation, modulated by the free-vortex magnitude $v_{\theta}(\rho_{i}^{t})$, intensifies as the agent approaches the drain and provides exploratory diversity in directions perpendicular to the radial pull. The random phase $\omega_{i}^{t}$ ensures that the spiral trajectory of every agent is distinct, which prevents the population from collapsing into a single trajectory.

\paragraph{Core phase.}
When the agent is in the immediate neighbourhood of its assigned drain, it samples positions in a small Gaussian cloud centred on the drain:
\begin{equation}
\Phi_{\mathrm{core}}(\mathbf{V}_{i}^{t}) = \mathbf{V}_{i}^{t} + \dfrac{\sigma(t)}{\sqrt{D}}\,\boldsymbol{\eta}_{i}^{t},
\quad
\sigma(t) = \sigma_{0}\,\dfrac{\phi(t)}{2},
\label{eq:phase3}
\end{equation}
where $\sigma_{0}$ is the initial core radius and $\sigma(t)$ shrinks together with $\phi(t)$. The core phase performs local refinement around the drain and is responsible for the final accuracy of the algorithm. The division by $\sqrt{D}$ keeps the typical step length constant across dimensions.

\subsubsection{Stochastic Vortex Switching}
\label{subsubsec:svs}

After the assignment of Eq.~\eqref{eq:assignment_score}, every agent has a small probability $p_{\mathrm{switch}}$ of being reassigned to a different drain. The switching probability $p_{\mathrm{switch}}$ is a fixed hyper-parameter of the algorithm. When a switch occurs, the new drain is sampled from a categorical distribution that excludes the currently assigned drain and is proportional to the remaining drain probabilities:
\begin{equation}
k_{i}^{t,\mathrm{new}} \sim \mathrm{Categorical}\!\left(\tilde{\mathbf{P}}_{i}^{t}\right),
\quad
\tilde{P}_{i,k}^{t} = \dfrac{P_{k}^{t}\,\mathbb{I}[k \ne k_{i}^{t}]}{\sum_{j \ne k_{i}^{t}} P_{j}^{t}},
\label{eq:svs}
\end{equation}
where $\mathbb{I}[\cdot]$ is the indicator function. Equation~\eqref{eq:svs} guarantees that a switch produces a real reassignment to a different drain, instead of returning the same drain with non-zero probability. After the switch, the assigned distance and the assigned drain position are recomputed for the new drain, and the agent then proceeds with one of the three motion phases according to its updated normalized distance. This mechanism preserves population diversity over the course of the search and supports a controlled form of basin-to-basin information exchange. The contribution of stochastic vortex switching to the overall performance of the algorithm is examined in the ablation study of Section~\ref{sec:results}.

\subsubsection{Splash-Out via L\'{e}vy Flight}
\label{subsubsec:splash}

To prevent the population from stagnating around a drain that does not contain the global optimum, DVO uses a splash-out mechanism that re-launches stuck agents into a distant region of the search space. A stagnation counter $\tau_{i}$ records the number of consecutive iterations during which agent $i$ has been in the core phase without improving its objective value. The counter is reset to zero whenever the agent improves, leaves the core phase, or undergoes a splash-out. When $\tau_{i} \ge \tau_{\mathrm{stay}}$, the agent becomes eligible for splash-out and is splashed with probability $p_{\mathrm{splash}}$. The splash step is generated by Mantegna's algorithm for L\'{e}vy flight~\cite{mantegna1994fast,yang2010nature}:
\begin{equation}
\boldsymbol{\ell}_{i}^{t} = \dfrac{\mathbf{u}}{|\mathbf{v}|^{1/\beta_{L}}},
\quad
\mathbf{u} \sim \mathcal{N}(\mathbf{0}, \sigma_{u}^{2}\,\mathbf{I}_{D}),
\quad
\mathbf{v} \sim \mathcal{N}(\mathbf{0}, \mathbf{I}_{D}),
\label{eq:levy_step}
\end{equation}
where the scale parameter $\sigma_{u}$ is computed from the stable index $\beta_{L} \in (0, 2)$ as
\begin{equation}
\sigma_{u} = \left(\dfrac{\Gamma(1+\beta_{L})\,\sin(\pi\beta_{L}/2)}{\Gamma\!\left((1+\beta_{L})/2\right)\,\beta_{L}\,2^{(\beta_{L}-1)/2}}\right)^{1/\beta_{L}}.
\label{eq:levy_sigma}
\end{equation}
The agent is then placed at
\begin{equation}
\mathbf{X}_{i}^{t+1} = \mathbf{V}_{1}^{t} + \dfrac{L_{\mathrm{splash}}}{\sqrt{D}}\,\boldsymbol{\ell}_{i}^{t},
\label{eq:splash_position}
\end{equation}
where $L_{\mathrm{splash}}$ is a fraction of the search-space diameter and $\mathbf{V}_{1}^{t}$ is the best drain at iteration $t$. The heavy-tailed distribution of L\'{e}vy steps allows splashed agents to travel arbitrarily far from the centre, which gives the algorithm a non-vanishing chance of escaping any local basin even in the late stages of the search. Equation~\eqref{eq:splash_position} captures the physical analogy of a fluid parcel being ejected from the surface of a vortex and falling back into the basin at a randomly chosen location.

\subsubsection{Boundary Handling and Best-So-Far Update}
\label{subsubsec:boundary}

After the position update, every component of $\mathbf{X}_{i}^{t+1}$ is clipped to the search bounds:
\begin{equation}
\mathbf{X}_{i}^{t+1} \leftarrow \max\!\left(\mathbf{lb},\,\min\!\left(\mathbf{ub},\,\mathbf{X}_{i}^{t+1}\right)\right).
\label{eq:clip_bounds}
\end{equation}
The objective function is then evaluated for the updated population, the global best solution is updated if a better candidate is discovered, and the drain positions are recomputed through the elitist rule of Eq.~\eqref{eq:elitism_pool}. The process repeats until the maximum number of iterations $T$ is reached.

\subsection{Pseudo-code}
\label{subsec:pseudocode}
\begin{figure}[H]
  \centering
  \includegraphics[width=0.85\linewidth]{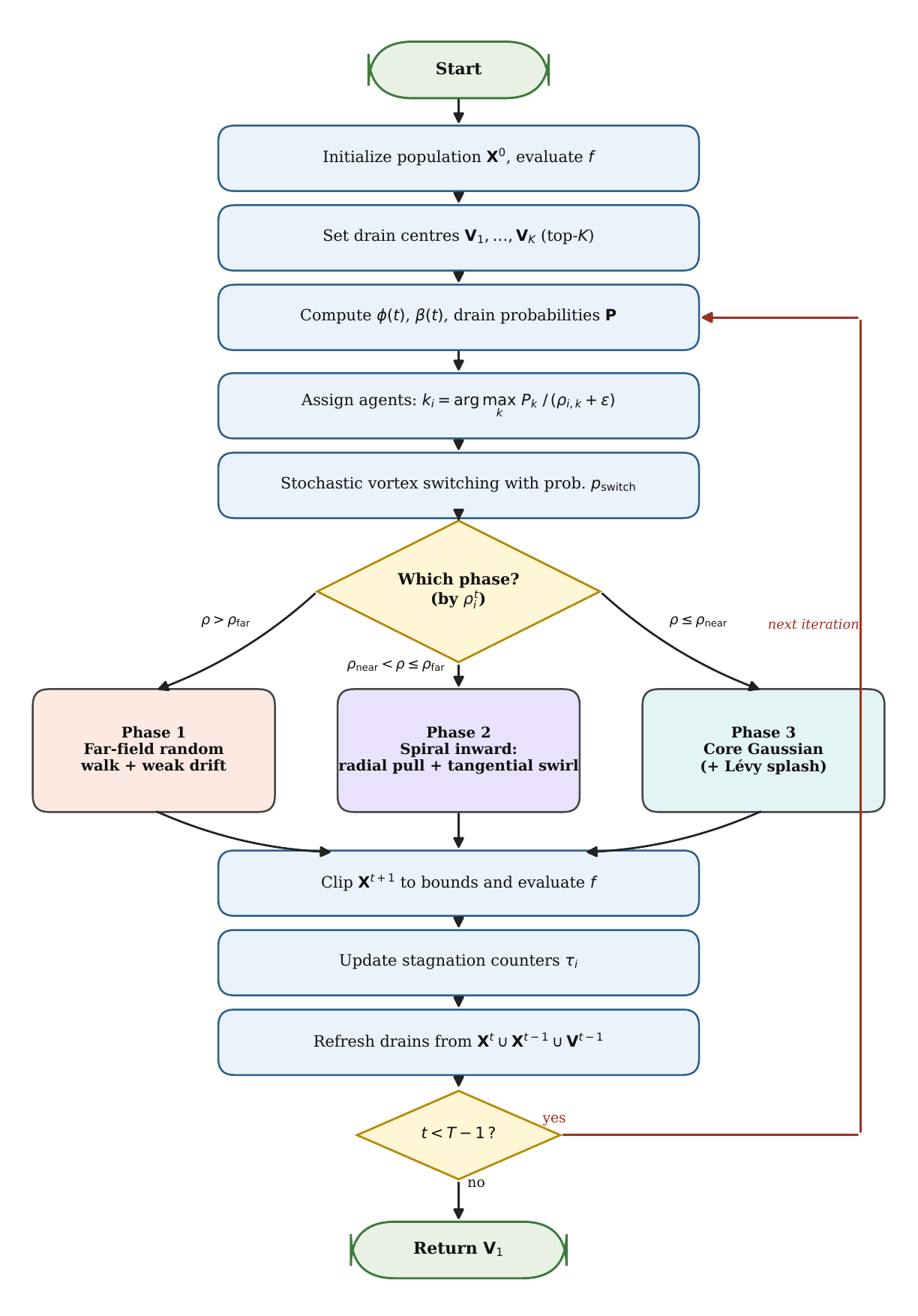}
  \caption{Flowchart of the DVO algorithm. Each iteration assigns every
  agent to a drain, applies optional stochastic vortex switching, and
  selects one of three motion phases according to the normalized distance
  $\rho_i^t$. The far-field, spiral, and core branches are colour-coded.
  The red path on the right indicates the loop-back to the next iteration.}
  \label{fig:dvo_flowchart}
\end{figure}

The full algorithm is summarized in Algorithm~\ref{alg:dvo}, and Fig.~\ref{fig:dvo_flowchart} illustrates the flow chart of the proposed algorithm. The algorithm takes as inputs the objective function $f$, the bounds $\mathbf{lb}$ and $\mathbf{ub}$, the population size $N$, the number of drains $K$, the maximum number of iterations $T$, and the hyper-parameters of the motion model. It returns the best solution found over the course of the search.

\begin{algorithm}[H]
\small
\caption{Drain-Vortex Optimization (DVO)}
\label{alg:dvo}
\begin{algorithmic}[1]
\REQUIRE Objective $f$, bounds $\mathbf{lb}, \mathbf{ub}$, population size $N$, drains $K$, iterations $T$, hyper-parameters $\{C, \gamma, c_{\min}, \alpha_{1}, \delta_{1}, \beta_{0}, \beta_{1}, \rho_{\mathrm{far}}, \rho_{\mathrm{near}}, \rho_{\mathrm{core}}, v_{\max}, \sigma_{0}, p_{\mathrm{switch}}, \tau_{\mathrm{stay}}, p_{\mathrm{splash}}, \beta_{L}, L_{\mathrm{splash}}\}$.
\ENSURE Best solution $\mathbf{V}_{1}$.
\STATE Initialize population $\mathbf{X}^{0}$ via Eq.~\eqref{eq:init} and evaluate $f(\mathbf{X}_{i}^{0})$ for all $i$.
\STATE Sort agents by fitness and set $\mathbf{V}_{1}^{0}, \dots, \mathbf{V}_{K}^{0}$ to the $K$ best solutions.
\STATE Initialize stagnation counters $\tau_{i} \leftarrow 0$ for all $i$.
\FOR{$t = 0, 1, \dots, T-1$}
    \STATE Compute $\phi(t)$ via Eq.~\eqref{eq:phi_schedule} and $\beta(t)$ via Eq.~\eqref{eq:beta_schedule}.
    \STATE Compute drain probabilities $\mathbf{P}^{t}$ via Eq.~\eqref{eq:drain_probability}.
    \STATE Compute distances $r_{i,k}^{t}$ and assign each agent to a drain via Eq.~\eqref{eq:assignment_score}.
    \STATE Apply stochastic vortex switching with probability $p_{\mathrm{switch}}$ via Eq.~\eqref{eq:svs}.
    \FOR{each agent $i = 1, \dots, N$}
        \IF{$\rho_{i}^{t} > \rho_{\mathrm{far}}$}
            \STATE Update $\mathbf{X}_{i}^{t+1}$ via the far-field rule, Eq.~\eqref{eq:phase1}.
        \ELSIF{$\rho_{i}^{t} > \rho_{\mathrm{near}}$}
            \STATE Update $\mathbf{X}_{i}^{t+1}$ via the spiral rule, Eqs.~\eqref{eq:vtheta}--\eqref{eq:phase2}.
        \ELSE
            \STATE Update $\mathbf{X}_{i}^{t+1}$ via the core rule, Eq.~\eqref{eq:phase3}.
            \IF{$\tau_{i} \ge \tau_{\mathrm{stay}}$ \AND $\mathrm{rand}(0,1) < p_{\mathrm{splash}}$}
                \STATE Re-launch agent via L\'{e}vy splash, Eqs.~\eqref{eq:levy_step}--\eqref{eq:splash_position}.
            \ENDIF
        \ENDIF
    \ENDFOR
    \STATE Clip $\mathbf{X}^{t+1}$ to the bounds via Eq.~\eqref{eq:clip_bounds} and evaluate $f$.
    \STATE Update stagnation counters $\tau_{i}$ based on improvement and current phase.
    \STATE Recompute drain positions $\mathbf{V}^{t+1}$ via the elitist pool of Eq.~\eqref{eq:elitism_pool}.
\ENDFOR
\RETURN $\mathbf{V}_{1}^{T}$
\end{algorithmic}
\end{algorithm}

\subsection{Computational Complexity}
\label{subsec:complexity}

The computational complexity of DVO at every iteration is dominated by three operations: the evaluation of the objective function for the $N$ agents, the computation of the pairwise distances between agents and drains, and the elitist drain selection from the candidate pool. The objective function evaluation contributes a problem-dependent term of $O(N \cdot \mathcal{C}_{f})$, where $\mathcal{C}_{f}$ is the cost of one objective evaluation. The distance computation involves $N \cdot K$ distances in $\mathbb{R}^{D}$ and contributes $O(N \cdot K \cdot D)$ operations. The elitist drain selection sorts a pool of at most $2N + K$ candidates and contributes $O((2N+K) \log(2N+K))$ operations. The remaining operations of the motion phases, the stochastic switching, and the splash-out are linear in $N \cdot D$ and do not change the overall scaling. Across $T$ iterations, the total complexity is therefore
\begin{equation}
O\big(T\,(N\,\mathcal{C}_{f} + N\,K\,D + N \log N)\big).
\label{eq:complexity}
\end{equation}
For typical settings with $K \ll N$ and $\mathcal{C}_{f}$ at least linear in $D$, the cost of the objective evaluation dominates the per-iteration cost, and DVO has the same asymptotic complexity as standard population-based metaheuristics such as Particle Swarm Optimization~\cite{kennedy1995particle}, Grey Wolf Optimizer~\cite{mirjalili2014grey}, and Whale Optimization Algorithm~\cite{mirjalili2016whale}. The memory complexity of DVO is $O((N+K)\,D)$ and is also comparable to standard methods.

\subsection{Discussion of Design Choices}
\label{subsec:design_choices}

Three design choices distinguish DVO from existing physics-based and swarm-based metaheuristics. The first is the explicit decomposition of the motion of every agent into a radial component and a tangential component, with magnitudes derived from the free-vortex law. Most existing methods couple position updates and velocity updates through a single rule that mixes attraction and perturbation. In DVO, the radial pull and the tangential rotation are independent terms that can be tuned separately. The second design choice is the adaptive spiral schedule of Eq.~\eqref{eq:shrink_radial}. The schedule maintains a non-zero radial pressure throughout the search and accelerates convergence in the final stages without removing the rotational diversity of the spiral phase. The third design choice is the population-level vortex basin assignment of Eq.~\eqref{eq:assignment_score}. The assignment combines drain quality and proximity, and ensures that the population is naturally distributed across several basins of attraction rather than collapsing into a single attractor. The optional stochastic vortex switching of Eq.~\eqref{eq:svs} adds a controlled form of basin-to-basin information exchange, and its specific contribution is studied empirically in Section~\ref{sec:results}.

%% file: sections/section_5_experimental_setup.tex
% =====================================================================
% Section 5: Experimental Setup
% =====================================================================

\section{Experimental Setup}
\label{sec:experimental_setup}

This section describes the experimental protocol used to evaluate DVO. Section~\ref{subsec:benchmark_suites} describes the benchmark suites. Section~\ref{sec:baselines} introduces the baseline algorithms and their parameter settings. Section~\ref{subsec:protocol} reports the experimental protocol, including the calibration policy and the stopping criterion. Section~\ref{subsec:metrics} defines the performance metrics and the statistical tests used in the analysis. Section~\ref{subsec:hardware} describes the hardware and the software environment.

\subsection{Benchmark Suites}
\label{subsec:benchmark_suites}

DVO is evaluated on four benchmark suites that together cover a broad spectrum of optimization landscapes. The first suite is the IEEE CEC 2022 special session and competition on single-objective bound-constrained numerical optimization~\cite{kumar2022cec2022}. The suite contains twelve test functions in dimension $D \in \{10, 20\}$, including unimodal, multimodal, hybrid, and composition functions. The second suite is the IEEE CEC 2017 special session~\cite{awad2017cec2017} on the same family of problems. Twenty-nine functions of CEC 2017 are used, namely $F_{1}$ and $F_{3}$ to $F_{30}$. Function $F_{2}$ is excluded as it has been reported as unstable and is recommended to be omitted in modern evaluations~\cite{awad2017cec2017}. The functions are evaluated in dimension $D \in \{30, 50\}$, which produces $29 \times 2 = 58$ benchmark cases.

The third suite is a collection of classical scalable and fixed-dimensional functions that are widely used in the metaheuristic literature~\cite{mirjalili2014grey,mirjalili2016whale,heidari2019harris}. The scalable subset contains thirteen functions $F_{1}$ to $F_{13}$, evaluated in dimension $D \in \{30, 100\}$. The fixed-dimensional subset contains ten functions $F_{14}$ to $F_{23}$ with their original (non-scalable) dimensions in the range $2$ to $6$. These two subsets are reported separately because they characterize different aspects of the algorithms. Scalable functions probe the dimensional behaviour of the optimizers, while fixed-dimensional functions probe their performance on small but highly multimodal landscapes.

The fourth suite consists of five constrained engineering design problems that are standard benchmarks for metaheuristic algorithms with constraint handling: three-bar truss, tension/compression spring, welded beam, pressure vessel, and speed reducer. The problems are taken from~\cite{coello2002theoretical,kannan1994augmented,sandgren1990nonlinear} and are evaluated with a static penalty constraint-handling method~\cite{coello2002theoretical}. The problem dimensions are determined by the original problem definitions and are not modified.

The role of each suite within the experimental protocol is summarized in Table~\ref{tab:experimental_protocol}. CEC 2022 is used as the calibration suite for the parameter settings of DVO. The remaining suites, namely CEC 2017, the classical functions, and the engineering problems, serve as out-of-sample validation suites. This separation between calibration and validation is detailed in Section~\ref{subsec:protocol}.

\input{sections/section_5_2_baselines.tex}
%\subsection{Baseline Algorithms}
%\label{subsec:baselines}

%Six well-established metaheuristic algorithms are used as baselines for the comparison. The selection covers classical swarm-based, modern swarm-inspired, trigonometric, arithmetic-operator, and physics-based methods. Particle Swarm Optimization (PSO)~\cite{kennedy1995particle} represents the classical swarm-based class. Grey Wolf Optimizer (GWO)~\cite{mirjalili2014grey} and Whale Optimization Algorithm (WOA)~\cite{mirjalili2016whale} represent the modern swarm-inspired class. Sine Cosine Algorithm (SCA)~\cite{mirjalili2016sca} represents the trigonometric class. Arithmetic Optimization Algorithm (AOA)~\cite{abualigah2021arithmetic} represents the arithmetic-operator class. Equilibrium Optimizer (EO)~\cite{faramarzi2020equilibrium} represents the physics-based class. The six baselines together provide a representative sample of the most widely cited metaheuristic algorithms of the last decade.

%The parameter values of every algorithm follow the settings reported in their original papers, as summarized in Table~\ref{tab:algorithm_settings}. No baseline-specific tuning is performed in this work, in order to keep the comparison fair and to limit the bias that may arise from algorithm-specific calibration. The population size is fixed at $N = 30$ for every algorithm and benchmark, which is the most common choice in the metaheuristic literature.

\subsection{Experimental Protocol and Calibration Policy}
\label{subsec:protocol}

A clear separation between calibration and validation is maintained throughout the study. The parameters of DVO are selected once during a preliminary calibration phase on CEC 2022. Several candidate configurations are explored at this stage, including variants with different drain counts, switching probabilities, drain attraction strengths, and adaptive schedules. After the calibration phase, the parameter set is frozen and used without further modification across all remaining benchmark suites, including CEC 2017, the classical functions, and the engineering problems. This protocol avoids per-function, per-dimension, or per-suite parameter adaptation and allows the CEC 2017 and engineering results to serve as out-of-sample validation of the proposed method. The final DVO parameter values are listed in Table~\ref{tab:algorithm_settings}, and the calibration policy is documented in the footnote of the same table.

The stopping criterion is identical for every algorithm and every benchmark case. The number of iterations is set to $T = 1000$, the population size is $N = 30$, and the total budget is therefore $30\,000$ function evaluations per run. Each run starts from a different random seed, and a total of $30$ independent runs is executed for every (algorithm, function, dimension) triplet. This produces a sample size that is consistent with standard practice in the metaheuristic literature and is sufficient to apply non-parametric statistical tests with adequate power~\cite{derrac2011practical}. For the engineering design problems, the same budget and number of independent runs are used.

The boundary handling rule for every algorithm is the standard component-wise clipping to the search bounds. For DVO, this clipping is applied after each motion phase, as described in Eq.~\eqref{eq:clip_bounds}. For the baseline algorithms, the same clipping is applied following the convention of their original implementations.

\subsection{Performance Metrics and Statistical Tests}
\label{subsec:metrics}

For unconstrained benchmark functions, the primary metric is the error $e = f^{*}_{\mathrm{found}} - f^{*}_{\mathrm{true}}$, where $f^{*}_{\mathrm{found}}$ is the best objective value found in a run and $f^{*}_{\mathrm{true}}$ is the global optimum value of the function. The mean, median, best, and standard deviation of the error are reported across the $30$ independent runs of each (algorithm, function, dimension) triplet. The mean error is the primary criterion used to declare the best algorithm in each row of the result tables, with the best value highlighted in bold.

For constrained engineering design problems, the primary metric is the best feasible objective value across the $30$ runs. The mean feasible objective, the feasibility rate, and the mean maximum constraint violation are also reported. Infeasible runs are not considered when ranking algorithms, but the feasibility rate is reported separately to allow the reader to assess the constraint satisfaction behaviour of each algorithm. The reporting convention is summarized in Table~\ref{tab:reporting_convention}.

Algorithms are ranked on every benchmark case according to their mean error or, for engineering problems, according to their best feasible objective. The average rank of an algorithm across a benchmark suite is then computed as the arithmetic mean of its per-case ranks. The algorithm with the lowest average rank is the best on the suite. The number of cases on which an algorithm produces the best mean error is denoted as the win count and is reported alongside the average rank. The win count and the average rank are complementary metrics, since an algorithm with the lowest average rank is not necessarily the one with the highest number of wins.

Two non-parametric statistical tests are used to assess the significance of the differences between algorithms. The Friedman test~\cite{friedman1937use} is used as a global test that determines whether the average ranks of the compared algorithms are significantly different from each other. The Friedman statistic and its $p$-value are reported on the CEC 2017 and CEC 2022 suites separately. The Wilcoxon signed-rank test~\cite{wilcoxon1945individual} is used as a pairwise test that compares DVO to each of the baseline algorithms. To account for the multiple comparison setting, $p$-values are corrected with the Holm step-down procedure~\cite{holm1979simple,derrac2011practical}. A $p$-value below $0.05$ after Holm correction is considered statistically significant. The Wilcoxon test is applied to the mean log-error of each algorithm across the $30$ runs of every benchmark case, which provides a stable measure of central tendency that is robust to the wide range of objective values across the benchmark functions.

The convergence behaviour of every algorithm is also analysed at six checkpoints during the search, namely at iteration $50$, $100$, $200$, $400$, $700$, and $1000$. The mean error and the log-error are reported at every checkpoint, together with a stability proxy that measures the inter-run variability of the error. These quantities are reported on a representative subset of CEC 2017 functions in dimension $D = 50$ and provide a finer-grained view of the search dynamics than the final-iteration metrics alone.

The ablation study uses the same statistical framework. Eight functions of CEC 2017 are sampled to cover the four landscape categories of the benchmark, namely unimodal, simple multimodal, hybrid, and composition. Eight variants of DVO are compared, in which one component of the algorithm is disabled at a time, and the same Wilcoxon test with Holm correction is applied to the mean log-error of every variant against the full DVO. The variants and the results of the ablation are described in Section~\ref{sec:results}.

\subsection{Hardware and Software Environment}
\label{subsec:hardware}

All experiments are executed on a workstation equipped with two NVIDIA L40S GPUs (48 GB memory each), an Intel Xeon CPU, and 256 GB of system memory. The implementation of DVO and of all baseline algorithms is fully vectorized in PyTorch~\cite{paszke2019pytorch} and exploits the GPU to execute the $30$ independent runs of every benchmark case in parallel. The objective functions of CEC 2017, CEC 2022, and the classical suite are also implemented in vectorized form on the GPU. Random seeds are fixed before every run to ensure reproducibility, and double-precision arithmetic is used in every component of the implementation. The source code, baseline implementations, experimental scripts, processed
results, and statistical analysis scripts are publicly available in the project
repository.

%% file: sections/section_5_2_baselines.tex
\subsection{Baseline Algorithms}
\label{sec:baselines}

Seven well-established metaheuristic algorithms are used as baselines
for the comparison. The selection covers classical swarm-based,
modern swarm-inspired, trigonometric, arithmetic-operator, and
physics-based methods. Particle Swarm Optimization
(PSO)~\cite{kennedy1995particle} represents the classical swarm-based
class. Grey Wolf Optimizer (GWO)~\cite{mirjalili2014grey} and Whale
Optimization Algorithm (WOA)~\cite{mirjalili2016whale} represent the
modern swarm-inspired class. Sine Cosine Algorithm
(SCA)~\cite{mirjalili2016sca} represents the trigonometric class.
Arithmetic Optimization Algorithm
(AOA)~\cite{abualigah2021arithmetic} represents the arithmetic-operator
class. Equilibrium Optimizer
(EO)~\cite{faramarzi2020equilibrium} represents the chemistry-inspired
physics-based class. Stratospheric Vortex Optimization Algorithm
(SVOA)~\cite{he2026stratospheric} represents the
atmospheric-flow physics-based class and is included because it shares
the high-level idea of building a search algorithm on a fluid-flow
analogy, while using a different physical model from the proposed
method. The seven baselines together provide a representative sample
of the most widely cited metaheuristic algorithms of the recent
literature.

The parameter values of every algorithm follow the settings reported
in their original papers, as summarized in
Table~\ref{tab:algorithm_settings}. No baseline-specific tuning is
performed in this work, in order to keep the comparison fair and to
limit the bias that may arise from algorithm-specific calibration.
The population size is fixed at $N=30$ for every algorithm and
benchmark, which is the most common choice in the metaheuristic
literature.

%% file: tables/table_experimental_protocol.tex
\begin{sidewaystable}
\centering
\caption{Experimental protocol used for evaluating DVO and baseline optimizers.}
\label{tab:experimental_protocol}
\footnotesize
\renewcommand{\arraystretch}{1.20}
\setlength{\tabcolsep}{4pt}
\resizebox{0.95\textheight}{!}{%
\begin{tabularx}{\textheight}{@{}l l X c c X@{}}
\toprule
\textbf{Experiment} & \textbf{Benchmark suite} & \textbf{Functions / problems} & \textbf{Dim.} & \textbf{Runs} & \textbf{Budget and metrics} \\
\midrule
Main benchmark I
  & CEC 2022 & F1--F12 & 10, 20 & 30
  & 30 particles, 1000 iterations, 30000 FEs; mean, median, best, std, rank. \\
Main benchmark II
  & CEC 2017 & F1, F3--F30 & 30, 50 & 30
  & 30 particles, 1000 iterations, 30000 FEs; mean, median, best, std, rank. \\
Classical scalable
  & Classical functions & F1--F13 & 30, 100 & 30
  & Reported separately for $D=30$ and $D=100$ to avoid mixing scalability effects. \\
Classical fixed-dim.
  & Classical functions & F14--F23 & 2--6 & 30
  & Reported separately from scalable functions. \\
Engineering design
  & Constrained engineering
  & Three-bar truss, spring, welded beam, pressure vessel, speed reducer
  & Specific & 30
  & Best feasible, mean feasible, feasibility rate, constraint violation. \\
Ablation study
  & CEC 2017 subset
  & F1, F3, F5, F10, F15, F20, F25, F30
  & 30, 50 & 30
  & Effect of greedy update, switching, vortex count, swirl, adaptive spiral, splash-out. \\
Convergence analysis
  & CEC 2017 subset & F10, F15, F30 & 50 & 30
  & Checkpoints at 50, 100, 200, 400, 700, 1000 iterations; mean error, log-error, stability proxy. \\
\bottomrule
\end{tabularx}}
\end{sidewaystable}

%% file: tables/table_algorithm_settings.tex
\begin{table*}[!t]
\centering
\caption{Parameter settings of DVO and the baseline optimizers.}
\label{tab:algorithm_settings}
\renewcommand{\arraystretch}{1.30}
\begin{tabularx}{\textwidth}{l X}
\toprule
\textbf{Algorithm} & \textbf{Parameter settings} \\
\midrule
DVO
  & $N=30$, $K=6$, $C=0.2$, $\gamma=0.5$, $p_{\mathrm{switch}}=0.08$,
    adaptive spiral enabled, greedy update enabled, forced splash
    replacement disabled. \\
PSO
  & $N=30$, $w$ linearly decreases from 0.9 to 0.4, $c_1=2.0$, $c_2=2.0$,
    $v_{\max}=0.2(\mathbf{ub}-\mathbf{lb})$. \\
GWO
  & $N=30$, convergence coefficient $a$ linearly decreases from 2 to 0. \\
WOA
  & $N=30$, spiral constant $b=1.0$, coefficient $a$ decreases from
    2 to 0. \\
SCA
  & $N=30$, coefficient $a=2.0$, number of elites $=2$. \\
AOA
  & $N=30$, $C_1=2.0$, $C_2=6.0$, $C_3=2.0$, $C_4=0.5$, acceleration
    range $[0.1,\,0.9]$. \\
EO
  & $N=30$, $a_1=2.0$, $a_2=1.0$, generation probability $GP=0.5$. \\
SVOA
  & $N=30$, baseline algorithm with reference parameter values from the
    original paper. \\
\bottomrule
\end{tabularx}
\end{table*}

%% file: tables/table_reporting_convention.tex
\begin{table}[!t]
\centering
\caption{Reporting convention used in the result tables.}
\label{tab:reporting_convention}
\renewcommand{\arraystretch}{1.25}
\begin{tabularx}{\linewidth}{l X}
\toprule
\textbf{Quantity} & \textbf{Rule} \\
\midrule
Unconstrained benchmark functions
  & Values are reported as the mean $\log_{10}$ of the absolute error
    over 30 independent runs. Lower values are better. The best value in
    each row is shown in \textbf{bold}. \\
Engineering design problems
  & Values are reported as the best feasible objective over 30 runs;
    the feasibility rate is shown in parentheses when it is below 1.00.
    Lower values are better. The best value in each row is shown in
    \textbf{bold}. \\
Average rank
  & Lower rank is better. The best average rank in each suite is shown in
    \textbf{bold}. \\
Statistical tests
  & The Friedman test is used for the global comparison across all
    algorithms. The Wilcoxon signed-rank test with Holm correction is
    used for the pairwise comparisons of DVO against each baseline. \\
\bottomrule
\end{tabularx}
\end{table}

%% file: sections/section_6_results.tex
\section{Results and Discussion}
\label{sec:results}

This section presents the experimental results obtained on the benchmark
suites described in Section~\ref{sec:experimental_setup}. The discussion
follows the same structure as the experimental protocol.
Section~\ref{sec:results_cec2022} reports the CEC 2022 calibration results.
Section~\ref{sec:results_cec2017} reports the out-of-sample validation on
CEC 2017. Section~\ref{sec:results_classical} discusses the classical
benchmark functions.
Section~\ref{sec:results_engineering} reports the constrained engineering
design problems. Section~\ref{sec:results_convergence} examines convergence
and stability on a representative CEC 2017 subset.
Section~\ref{sec:results_ablation} reports the ablation study, and
Section~\ref{sec:results_statistics} summarizes the main statistical
evidence.

All numerical values reported in this section are mean $\log_{10}$ errors
over 30 independent runs, unless stated otherwise. The use of the
$\log_{10}$ scale removes the wide dynamic range of the raw mean errors
across the benchmark functions and makes the per-row comparisons easier
to read. Engineering design results are reported as best feasible
objective values and are not transformed to the log scale.

\subsection{CEC 2022 (Calibration Suite)}
\label{sec:results_cec2022}

The CEC 2022 suite contains twelve test functions evaluated in two
dimensions, $D \in \{10, 20\}$, which gives twenty-four benchmark cases
in total. The detailed results are reported in
Table~\ref{tab:cec2022}. For each case, the table reports the
mean $\log_{10}$ error obtained over 30 independent runs, and the best
mean value in each row is highlighted.

DVO obtains the best mean log-error on 8 of the 24 cases. PSO obtains 9
wins and SVOA obtains 7 wins on this suite. GWO, WOA, SCA, AOA, and EO
do not obtain the best mean log-error on any CEC 2022 case. The average
rank of DVO is 2.13, followed by PSO at 2.44, SVOA at 2.52, EO at 3.88,
GWO at 4.54, SCA at 5.98, AOA at 6.60, and WOA at 7.92.

The Friedman test shows a clear difference among the algorithms on this
suite, with a statistic of 130.07 and a $p$-value of
$6.08 \times 10^{-25}$. The Wilcoxon signed-rank test with Holm
correction shows that DVO is significantly better than WOA, SCA, AOA,
GWO, and EO at the 5\% level. The differences against PSO and SVOA are
not significant. The mean log-error of DVO is 1.50, while the mean
log-errors of PSO and SVOA are 0.89 and 1.47, respectively. The CEC 2022
results should not be read as a complete dominance of DVO over PSO or
SVOA on this suite. Instead, they show that DVO is highly competitive
on the calibration suite and obtains the best overall rank across the
twenty-four cases, with PSO and SVOA forming a closely competing group.

It is important to interpret this suite carefully. The DVO parameter
values were selected during a preliminary calibration phase on CEC 2022,
as described in Section~\ref{sec:experimental_setup}. Therefore, CEC
2022 is used to fix the final configuration of DVO, not to provide fully
independent validation. The stronger test of generalization is the
evaluation of the same fixed configuration on CEC 2017, the classical
suite, and the engineering design problems.

\subsection{CEC 2017 (Out-of-Sample Validation)}
\label{sec:results_cec2017}

The CEC 2017 suite is the largest benchmark in this study. It contains
twenty-nine functions, namely F1 and F3 to F30, evaluated in two
dimensions, $D \in \{30, 50\}$. This gives fifty-eight benchmark cases.
The detailed results for $D = 30$ and $D = 50$ are reported in
Table~\ref{tab:cec2017_d30} and
Table~\ref{tab:cec2017_d50}, respectively.

The CEC 2017 results provide the main out-of-sample evidence for the
proposed method. DVO obtains the best mean log-error on 34 of the 58
cases, which corresponds to a 59\% win rate. PSO obtains 15 wins, SVOA
obtains 8 wins, and EO obtains 1 win. GWO, WOA, SCA, and AOA do not
obtain any best mean log-error on this suite. DVO also achieves the
best average rank, with an average rank of 1.66. The next algorithms
are PSO (2.42), SVOA (2.74), EO (3.93), GWO (4.29), AOA (6.11),
SCA (6.92), and WOA (7.91).

These results are meaningful because no additional tuning was performed
on CEC 2017. The same parameter set calibrated on CEC 2022 was used
without modification. CEC 2017 is also more demanding in this study
because it is larger and uses higher dimensions than the calibration
suite. The strong performance on this suite supports the generalization
ability of the drain-vortex motion model on shifted, rotated, hybrid,
and composition landscapes.

The Friedman test on the CEC 2017 ranks returns a statistic of 347.44
with a $p$-value of $4.36 \times 10^{-71}$, which indicates a strong
difference among the compared algorithms. The Wilcoxon signed-rank test
with Holm correction shows that DVO is significantly better than every
baseline at the 5\% level on the combined CEC 2017 suite. The
Holm-corrected $p$-values are below $10^{-9}$ for WOA, SCA, AOA, and
GWO, below $10^{-8}$ for EO, below $10^{-2}$ for PSO, and equal to
$6.08 \times 10^{-3}$ for SVOA. The mean log-error gap between DVO and
the baselines ranges from $-0.05$ against SVOA to $-1.39$ against WOA.
The detailed pairwise statistics are reported in Table~\ref{tab:wilcoxon}.

Overall, CEC 2017 is the strongest evidence in favour of DVO in this
paper. It shows that the calibrated DVO configuration is not only
competitive on the calibration suite, but also transfers well to a
larger and more difficult benchmark.

\subsection{Classical Benchmark Functions}
\label{sec:results_classical}

The classical benchmark suite contains thirteen scalable functions,
F1 to F13, evaluated in $D \in \{30, 100\}$, and ten fixed-dimensional
functions, F14 to F23. The results are reported separately for the
three subsets. Table~\ref{tab:classical_d30} reports the
scalable functions in $D = 30$, Table~\ref{tab:classical_d100}
reports the scalable functions in $D = 100$, and
Table~\ref{tab:classical_fixed} reports the fixed-dimensional
functions.

The results on the scalable classical functions are mixed. WOA performs
best on this subset. It obtains the best mean log-error on 9 of the 13
cases in $D = 30$ and on 12 of the 13 cases in $D = 100$. Its average
ranks are 1.69 and 1.46, respectively. EO and GWO also perform strongly
on these functions. DVO is mid-table, with an average rank of 6.08 in
$D = 30$ and 5.39 in $D = 100$. SVOA is also mid-table, with average
ranks of 6.42 and 6.23, respectively. These results show that DVO is
not the best choice for every type of benchmark landscape, and the
same is true for SVOA.

This behaviour can be explained by the structure of the classical
scalable functions. Many of them have a global optimum at the origin
and have weak or no coupling between dimensions. On such landscapes,
algorithms that contract quickly toward a single attractor can be very
effective. WOA, EO, and GWO benefit from this structure and often reach
extremely small errors. DVO and SVOA, by contrast, were designed to
maintain structured diversity through rotational or basin-based
mechanisms, which is useful on coupled landscapes but less efficient on
simple separable problems.

The fixed-dimensional subset gives a different picture. These functions
are low-dimensional but highly multimodal. PSO obtains the best
performance, with 6 wins and an average rank of 1.65. SVOA is the
second-best on this subset, with 5 wins and an average rank of 2.15.
DVO ranks fifth, with an average rank of 4.70, behind EO (3.00) and
GWO (4.30). DVO does not obtain the best mean log-error on any
fixed-dimensional case. This indicates that the drain-vortex motion is
less effective on small fixed-dimensional multimodal landscapes than on
the larger CEC 2017 cases.

The classical benchmark results are useful because they define the
operating regime of DVO more clearly. DVO performs strongly on modern
shifted, rotated, and heterogeneous competition suites, but it is less
effective on simple separable scalable functions where aggressive
contraction is enough, and on small fixed-dimensional multimodal
problems where PSO and SVOA dominate. This limitation is reported
explicitly to avoid overclaiming and to provide a balanced
interpretation of the algorithm.

\subsection{Engineering Design Problems}
\label{sec:results_engineering}

The engineering benchmark suite contains five constrained design
problems: three-bar truss, tension/compression spring, welded beam,
pressure vessel, and speed reducer. The detailed results are reported
in Table~\ref{tab:engineering}. The table reports the best
feasible objective value, with the feasibility rate shown in parentheses
when it is below 1.00.

On this suite, PSO obtains the best average rank (2.30) and the best
feasible objective on 4 of the 5 problems. EO and GWO follow, with
average ranks of 2.50 and 3.10. AOA ranks fourth, with an average rank
of 4.50, followed by DVO (5.10) and SVOA (5.70). DVO does not obtain
the best feasible objective on any engineering problem. The feasibility
rate of DVO is 100\% on four of the five problems and 90\% on the
speed reducer problem, which gives an overall feasibility rate of 98\%
across the suite. SVOA also produces feasible solutions in nearly every
run, with a feasibility rate below 1.00 only on the pressure vessel
problem (97\%).

The engineering results suggest that the current DVO configuration is
not specialized for small constrained problems. The engineering
problems used here have low dimensionality, usually between 2 and 7,
and the quality of the result depends strongly on constraint handling.
In this study, DVO uses the same static penalty approach as the
baselines and does not include a dedicated feasibility-preserving
mechanism. In addition, the multi-vortex structure distributes the
population over several basins, which is useful on large heterogeneous
landscapes but less efficient when the feasible region is small and
the dimension is low.

The absolute differences should also be considered. On the three-bar
truss problem, all algorithms reach the same best feasible objective
of 263.9. On the welded beam, DVO obtains 1.735, while PSO obtains
1.725. On the speed reducer, DVO obtains 2997, while the best baseline
obtains 2994. These differences affect the rank-based metrics, but they
are relatively small in practical terms. The engineering suite
therefore shows that DVO and SVOA are feasible and competitive, but not
superior to the strongest baselines under the current penalty-based
constraint-handling setting.

\subsection{Convergence Analysis}
\label{sec:results_convergence}

\begin{figure*}[!t]
  \centering
  \begin{subfigure}[b]{0.32\linewidth}
    \centering
    \includegraphics[width=\linewidth]{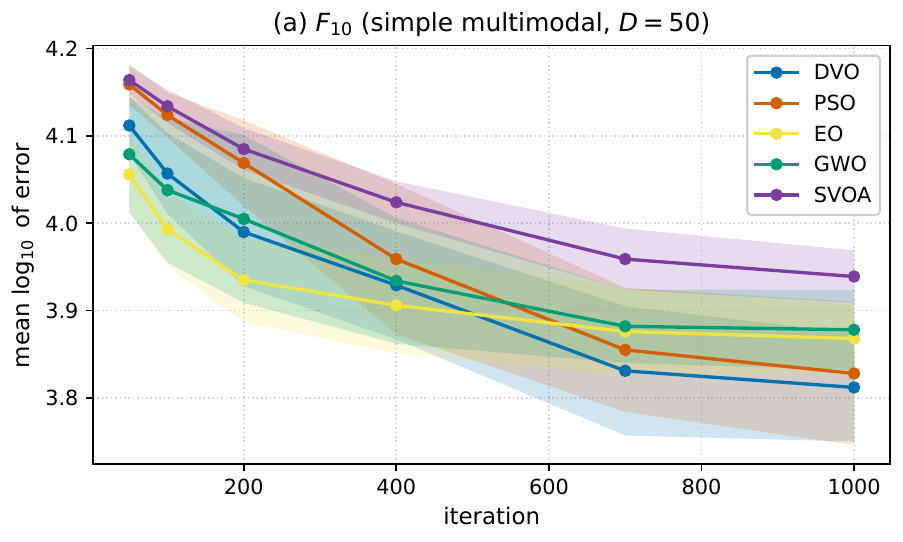}
    \caption{$F_{10}$ (simple multimodal), $D=50$}
    \label{fig:conv_F10}
  \end{subfigure}
  \hfill
  \begin{subfigure}[b]{0.32\linewidth}
    \centering
    \includegraphics[width=\linewidth]{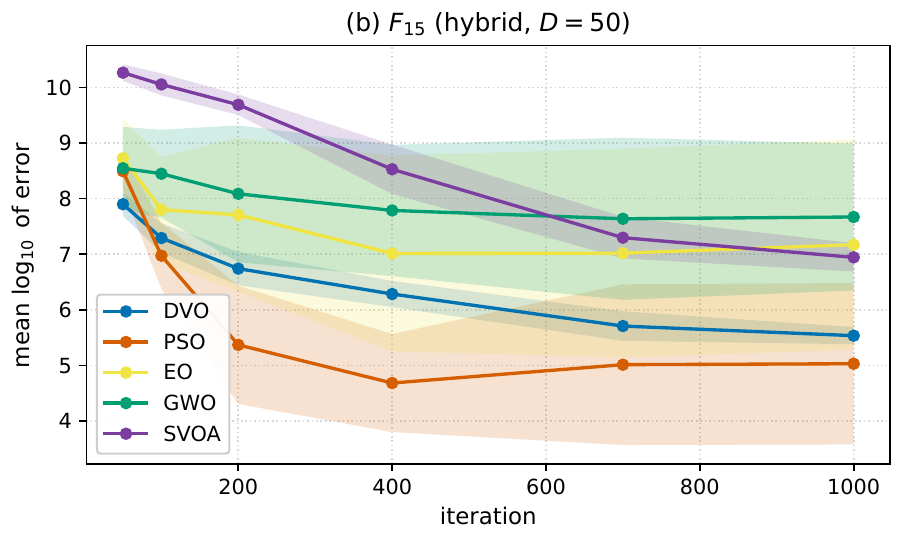}
    \caption{$F_{15}$ (hybrid), $D=50$}
    \label{fig:conv_F15}
  \end{subfigure}
  \hfill
  \begin{subfigure}[b]{0.32\linewidth}
    \centering
    \includegraphics[width=\linewidth]{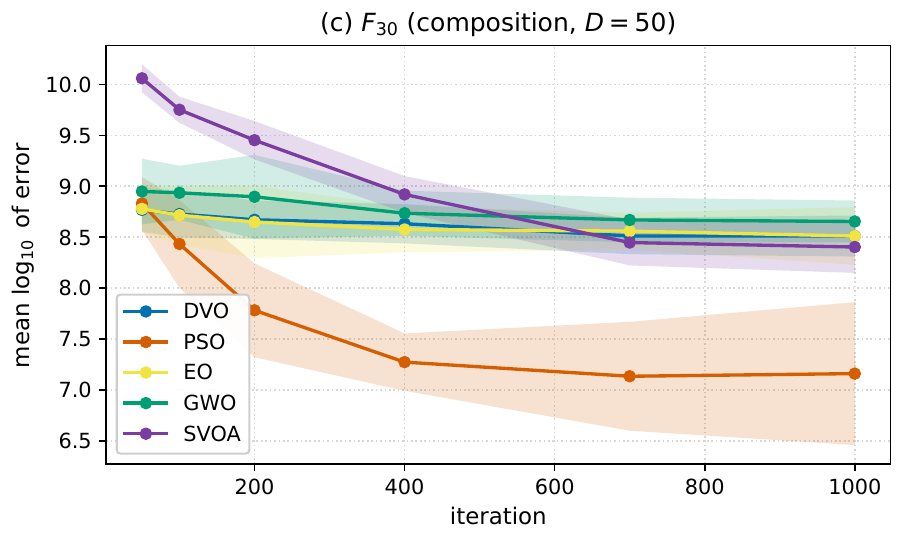}
    \caption{$F_{30}$ (composition), $D=50$}
    \label{fig:conv_F30}
  \end{subfigure}
  \caption{Convergence curves of DVO and the strongest baselines on the
  CEC 2017 convergence subset. Lower mean $\log_{10}$ error indicates
  better performance. Shaded bands show inter-run variability.}
  \label{fig:convergence}
\end{figure*}

\begin{figure*}[!t]
  \centering
  \begin{subfigure}[b]{0.32\linewidth}
    \centering
    \includegraphics[width=\linewidth]{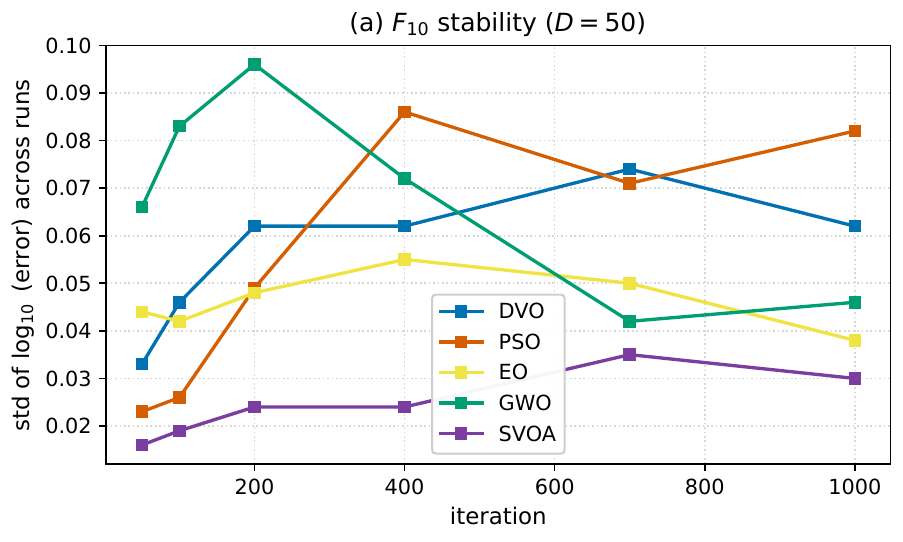}
    \caption{$F_{10}$, $D=50$}
    \label{fig:stab_F10}
  \end{subfigure}
  \hfill
  \begin{subfigure}[b]{0.32\linewidth}
    \centering
    \includegraphics[width=\linewidth]{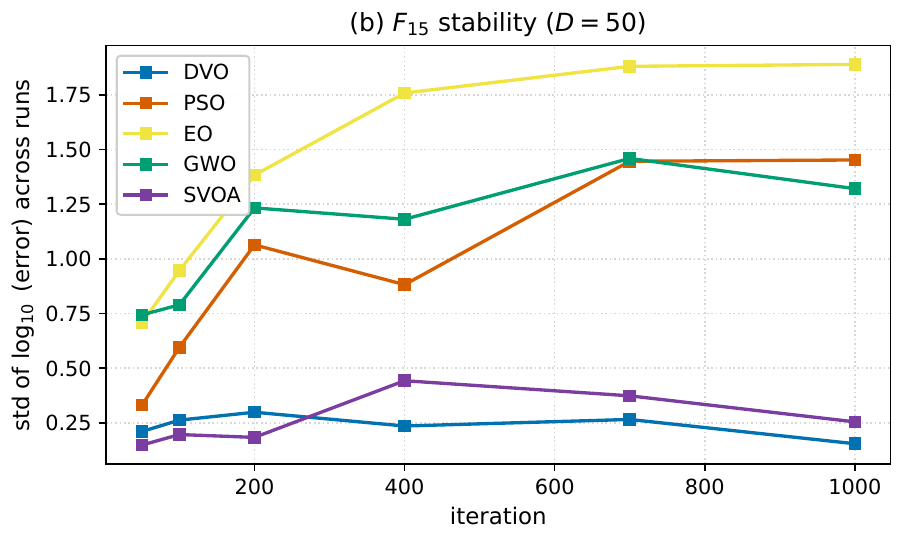}
    \caption{$F_{15}$, $D=50$}
    \label{fig:stab_F15}
  \end{subfigure}
  \hfill
  \begin{subfigure}[b]{0.32\linewidth}
    \centering
    \includegraphics[width=\linewidth]{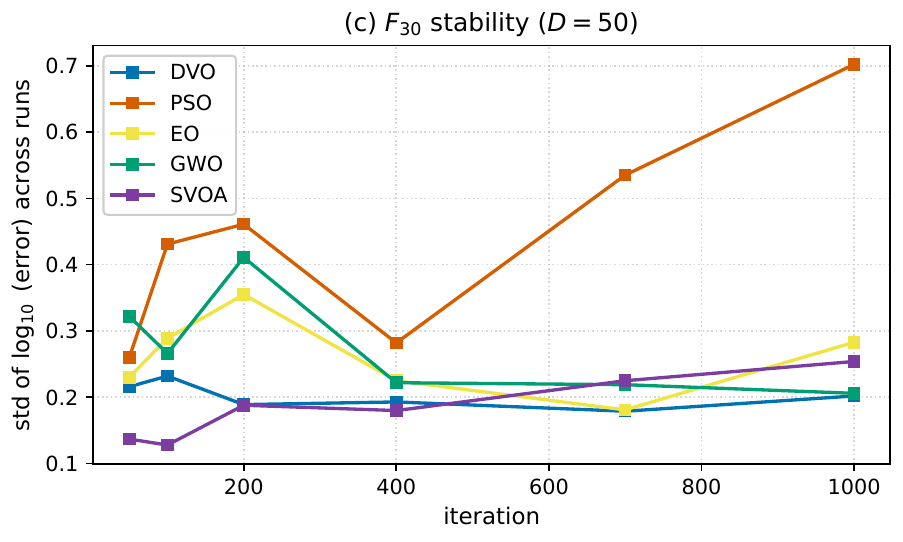}
    \caption{$F_{30}$, $D=50$}
    \label{fig:stab_F30}
  \end{subfigure}
  \caption{Stability profiles: standard deviation of $\log_{10}$ error
  across the 30 independent runs at each checkpoint, on the CEC 2017
  convergence subset. Lower values indicate more reproducible search
  behaviour.}
  \label{fig:stability}
\end{figure*}

The convergence behaviour of DVO is examined on a representative subset
of CEC 2017 in dimension $D = 50$. The subset contains three functions
from different categories: $F_{10}$, a simple multimodal function;
$F_{15}$, a hybrid function; and $F_{30}$, a composition function. DVO is compared with PSO, GWO, EO, and SVOA on this subset. These
algorithms are selected because they represent the strongest competitors in
the main numerical comparisons and because SVOA is the closest recent
fluid-flow-inspired baseline. The convergence curves are shown in
Fig.~\ref{fig:convergence}, and the corresponding stability profiles
are shown in Fig.~\ref{fig:stability}.

% Convergence figure (Fig. fig:convergence)

% Stability figure (Fig. fig:stability)

The convergence subset is restricted to PSO, GWO, and EO so that the
convergence trajectories remain readable in the figures. SVOA is
excluded from the convergence panel because the per-suite results
already show that SVOA and PSO occupy a similar performance range on
CEC 2017, and adding a fifth curve does not change the qualitative
picture but reduces the visual clarity of the comparison.

On $F_{10}$, DVO obtains the lowest final mean log-error, followed by
PSO, EO, and GWO. The standard deviation of DVO is also among the
lowest, which indicates a stable convergence pattern. On $F_{15}$, PSO
obtains the lowest mean log-error, while DVO obtains a higher value but
with a much smaller standard deviation. On $F_{30}$, PSO obtains the
lowest mean log-error, followed by DVO and EO with similar values, and
GWO above them. DVO has the smallest standard deviation among the compared algorithms on $F_{30}$.

The convergence analysis therefore shows a trade-off. PSO can be
faster or more aggressive on some functions, especially $F_{15}$ and
$F_{30}$. DVO, however, produces more stable trajectories across
independent runs. This stability is consistent with the multi-stage
structure of the algorithm. The far-field, spiral, and core phases
reduce abrupt collapse, while the drain assignment mechanism keeps the
population organized around promising regions. This behaviour is
useful when repeatability is important and when the user does not want
the final result to depend strongly on a favourable random seed.

\subsection{Ablation Study}
\label{sec:results_ablation}

The ablation study examines the contribution of the main components of
DVO. Eight CEC 2017 functions are selected to cover the four landscape
categories of the suite: unimodal (F1, F3), simple multimodal (F5,
F10), hybrid (F15, F20), and composition (F25, F30). Each function is
evaluated in $D \in \{30, 50\}$, giving 16 cases per variant. The
ablation results focus on the internal components of DVO and are
independent of the choice of baselines. They are therefore reported in
the same form as in the original calibration phase, without the
addition of SVOA, and the conclusions are unchanged with respect to
the previous version of the paper. The detailed results are reported
in Table~\ref{tab:ablation_cec2017_subset_results}.

Eight variants are compared. The full DVO includes all components.
The variant \texttt{no\_adaptive\_spiral} removes the adaptive spiral
schedule. The variant \texttt{no\_swirl} removes the tangential
component. The variant \texttt{no\_switch} disables stochastic vortex
switching. The variant \texttt{no\_splash} disables the splash-out
mechanism. The variant \texttt{no\_greedy} disables the greedy agent
update. The variant \texttt{radial\_only} keeps only the radial
component of the spiral phase. Finally, \texttt{single\_vortex} sets
$K=1$ and removes the multi-vortex structure.

The ablation results should be interpreted cautiously because the
subset contains only 16 cases. The Friedman test indicates that the
variants do not behave identically on the overall subset. However,
after Holm correction, none of the pairwise Wilcoxon comparisons
between the full DVO and an ablated variant is significant at the 5\%
level. The smallest unadjusted $p$-value is obtained for
\texttt{no\_adaptive\_spiral}, but it is not significant after
correction. This means that the ablation subset is not large enough
to prove the isolated contribution of each component statistically.

Even so, the ranking pattern is informative. The worst average ranks
are obtained by \texttt{no\_adaptive\_spiral} and \texttt{no\_swirl}.
This suggests that the adaptive spiral schedule and the tangential
free-vortex component are important to the search dynamics. The full
DVO ranks in the middle of the ablation group, while
\texttt{single\_vortex} and \texttt{no\_switch} perform competitively
on this limited subset. This observation indicates that the
drain-vortex radial--tangential motion is the core mechanism of the
method, while the multi-vortex structure acts mainly as a diversity
mechanism whose value depends on the benchmark class.

This interpretation is consistent with the full CEC 2017 results. The
calibrated multi-vortex configuration obtains the best aggregate
performance on the complete CEC 2017 suite, with 34 wins in 58 cases.
However, the ablation subset shows that not every component improves
every problem class in isolation. The components are therefore better
understood as a combined search design rather than as independently
dominant operators.

\subsection{Statistical Analysis Across Suites}
\label{sec:results_statistics}

\begin{figure*}[!t]
  \centering
  \includegraphics[width=0.95\linewidth]{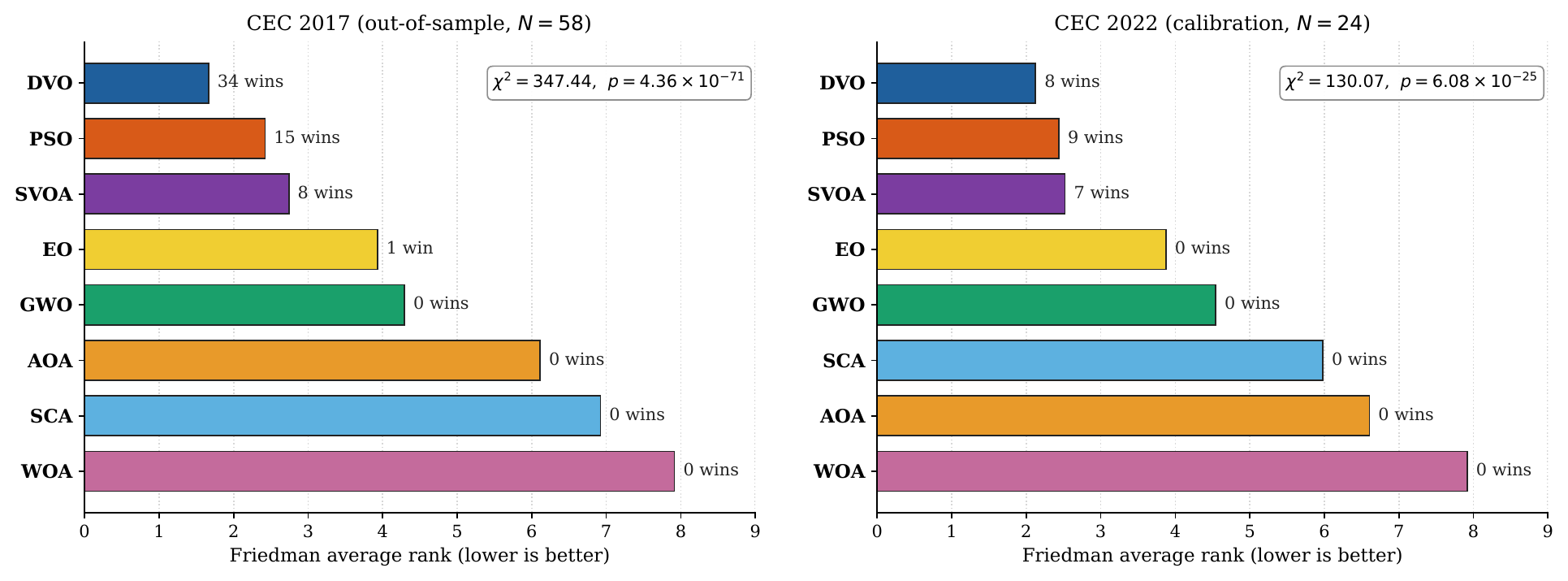}
  \caption{Average Friedman ranks of DVO and the seven baseline
  algorithms on CEC 2017 (out-of-sample, $N=58$) and CEC 2022
  (calibration, $N=24$). Lower rank indicates better performance. Win
  counts are shown next to each bar. The Friedman statistic and its
  $p$-value confirm that the differences among algorithms are not due
  to random variation alone.}
  \label{fig:friedman_ranks}
\end{figure*}

The non-parametric statistical tests provide a compact summary of the
main numerical evidence. On CEC 2022, the Friedman test gives a
statistic of 130.07 with a $p$-value of $6.08 \times 10^{-25}$. On
CEC 2017, the Friedman test gives a statistic of 347.44 with a
$p$-value of $4.36 \times 10^{-71}$. These results show that the
differences among algorithms are not due to random variation alone.
The detailed Friedman ranks are reported in Table~\ref{tab:friedman}
and visualized in Fig.~\ref{fig:friedman_ranks}.

% Friedman ranks bar chart (Fig. fig:friedman_ranks)

The Wilcoxon signed-rank test with Holm correction gives a more
detailed pairwise view, reported in Table~\ref{tab:wilcoxon}. On
CEC 2017, DVO is significantly better than every baseline, including
SVOA. On CEC 2022, DVO is significantly better than five of the seven
baselines (WOA, SCA, AOA, GWO, EO), while the differences against PSO
and SVOA are not significant. This is an important result because it
shows both the strength and the limitation of the proposed algorithm.
DVO is not claimed to dominate every baseline on every suite, but it
provides the best rank and a strong win count on the two modern
competition benchmarks considered in this study.

The overall picture is therefore balanced. DVO performs strongly on
CEC 2022 and CEC 2017, where the functions are shifted, rotated,
hybrid, or composite. It is less effective on the classical scalable
functions, where simple separable structures favour aggressive
single-attractor contraction. It is also not the leading method on the
small constrained engineering problems under the static penalty
setting used here. SVOA shows a similar profile on the modern
competition suites and a relatively better profile on the classical
fixed-dimensional functions, which is consistent with the design
philosophy of that algorithm. These results suggest that DVO is most
suitable for heterogeneous continuous optimization landscapes where
maintaining structured diversity is useful.

From an algorithmic perspective, the results support the main design
idea of DVO: a search process based on radial attraction, tangential
free-vortex motion, adaptive spiral exploitation, and population-level
basin assignment. The empirical evidence does not suggest that every
component is equally important in every problem class. Instead, it
suggests that the combination of these components gives a robust
search behaviour on modern, coupled benchmark landscapes. This is the
main practical conclusion drawn from the experimental study.

%% file: tables/table_cec2022_results.tex
\begin{table*}[!t]
\centering
\caption{CEC 2022 results. Values are mean $\log_{10}$ error over 30 independent runs. Lower values are better. The best value in each row is shown in \textbf{bold}.}
\label{tab:cec2022}
\scriptsize
\renewcommand{\arraystretch}{1.10}
\setlength{\tabcolsep}{4pt}
\begin{tabularx}{\textwidth}{@{}l >{\centering\arraybackslash}X >{\centering\arraybackslash}X >{\centering\arraybackslash}X >{\centering\arraybackslash}X >{\centering\arraybackslash}X >{\centering\arraybackslash}X >{\centering\arraybackslash}X >{\centering\arraybackslash}X@{}}
\toprule
\textbf{Case} & \textbf{DVO} & \textbf{SVOA} & \textbf{PSO} & \textbf{GWO} & \textbf{WOA} & \textbf{SCA} & \textbf{AOA} & \textbf{EO} \\
\midrule
F01 D10 & -1.100 & 1.897 & \textbf{-6.930} & 3.076 & 4.100 & 2.991 & 3.318 & 2.518 \\
F02 D10 & 0.625 & \textbf{0.434} & 1.194 & 1.322 & 2.237 & 1.813 & 1.544 & 0.886 \\
F03 D10 & -0.110 & 0.362 & \textbf{-6.602} & 0.345 & 1.625 & 1.287 & 1.373 & 0.042 \\
F04 D10 & \textbf{1.007} & 1.113 & 1.133 & 1.227 & 1.675 & 1.617 & 1.488 & 1.198 \\
F05 D10 & -1.730 & -1.288 & \textbf{-4.596} & 0.714 & 2.727 & 2.075 & 2.604 & 1.088 \\
F06 D10 & 3.214 & 3.605 & \textbf{3.203} & 3.457 & 3.814 & 6.382 & 3.863 & 3.304 \\
F07 D10 & 1.294 & 1.532 & \textbf{0.900} & 1.537 & 1.945 & 1.744 & 1.789 & 1.569 \\
F08 D10 & 1.309 & 1.423 & \textbf{0.988} & 1.430 & 1.588 & 1.533 & 1.465 & 1.365 \\
F09 D10 & 2.368 & \textbf{2.363} & 2.363 & 2.446 & 2.587 & 2.427 & 2.504 & 2.405 \\
F10 D10 & 2.112 & \textbf{2.002} & 2.151 & 2.149 & 2.559 & 2.050 & 2.194 & 2.137 \\
F11 D10 & 0.703 & \textbf{-6.187} & 0.685 & 2.137 & 2.607 & 2.246 & 2.387 & 1.842 \\
F12 D10 & \textbf{2.211} & 2.215 & 2.261 & 2.226 & 2.389 & 2.231 & 2.272 & 2.221 \\
F01 D20 & \textbf{1.159} & 3.988 & 3.060 & 4.021 & 4.656 & 4.211 & 4.240 & 4.052 \\
F02 D20 & \textbf{1.706} & 1.716 & 1.830 & 2.087 & 2.951 & 2.552 & 2.631 & 2.085 \\
F03 D20 & 0.703 & 1.141 & \textbf{-0.131} & 1.092 & 1.870 & 1.655 & 1.653 & 1.142 \\
F04 D20 & \textbf{1.511} & 1.696 & 1.621 & 1.753 & 2.208 & 2.153 & 2.056 & 1.782 \\
F05 D20 & 1.219 & \textbf{0.101} & 1.286 & 2.817 & 3.444 & 3.135 & 3.324 & 2.810 \\
F06 D20 & 4.010 & 4.530 & \textbf{3.738} & 4.812 & 8.202 & 7.954 & 5.959 & 4.251 \\
F07 D20 & 1.847 & 1.951 & \textbf{1.687} & 1.985 & 2.365 & 2.165 & 2.225 & 1.937 \\
F08 D20 & 1.759 & \textbf{1.529} & 1.743 & 1.630 & 2.031 & 1.826 & 1.934 & 1.683 \\
F09 D20 & \textbf{2.258} & 2.295 & 2.315 & 2.371 & 2.691 & 2.474 & 2.454 & 2.330 \\
F10 D20 & 3.060 & \textbf{2.054} & 2.499 & 3.014 & 3.471 & 2.514 & 2.868 & 2.786 \\
F11 D20 & \textbf{2.484} & 2.494 & 2.536 & 3.009 & 3.629 & 3.360 & 3.409 & 3.030 \\
F12 D20 & \textbf{2.408} & 2.409 & 2.493 & 2.473 & 2.728 & 2.571 & 2.571 & 2.444 \\
\bottomrule
\end{tabularx}
\end{table*}

%% file: tables/table_cec2017_D30_results.tex
\begin{table}[p]
\centering
\caption{CEC 2017 results for 30-dimensional functions. Values are mean $\log_{10}$ error over 30 independent runs. Lower values are better. The best value in each row is shown in \textbf{bold}.}
\label{tab:cec2017_d30}
\scriptsize
\renewcommand{\arraystretch}{1.10}
\setlength{\tabcolsep}{4pt}
\begin{tabularx}{\textwidth}{@{}l >{\centering\arraybackslash}X >{\centering\arraybackslash}X >{\centering\arraybackslash}X >{\centering\arraybackslash}X >{\centering\arraybackslash}X >{\centering\arraybackslash}X >{\centering\arraybackslash}X >{\centering\arraybackslash}X@{}}
\toprule
\textbf{Case} & \textbf{DVO} & \textbf{SVOA} & \textbf{PSO} & \textbf{GWO} & \textbf{WOA} & \textbf{SCA} & \textbf{AOA} & \textbf{EO} \\
\midrule
F01 D30 & 7.326 & \textbf{3.272} & 7.736 & 10.588 & 11.523 & 11.236 & 11.326 & 10.592 \\
F03 D30 & \textbf{3.594} & 4.791 & 4.568 & 4.672 & 5.296 & 4.807 & 4.806 & 4.661 \\
F04 D30 & \textbf{1.998} & 2.052 & 2.340 & 2.527 & 3.886 & 3.296 & 3.260 & 2.464 \\
F05 D30 & \textbf{1.764} & 2.009 & 1.949 & 2.194 & 2.589 & 2.487 & 2.449 & 2.179 \\
F06 D30 & \textbf{1.160} & 1.487 & 1.293 & 1.615 & 2.020 & 1.882 & 1.902 & 1.524 \\
F07 D30 & \textbf{2.097} & 2.296 & 2.116 & 2.385 & 2.854 & 2.705 & 2.776 & 2.352 \\
F08 D30 & \textbf{1.780} & 1.989 & 1.889 & 2.096 & 2.507 & 2.444 & 2.368 & 2.099 \\
F09 D30 & 2.087 & \textbf{1.693} & 2.415 & 3.383 & 3.978 & 3.764 & 3.849 & 3.365 \\
F10 D30 & \textbf{3.525} & 3.628 & 3.543 & 3.589 & 3.866 & 3.884 & 3.713 & 3.565 \\
F11 D30 & 2.304 & 2.734 & \textbf{2.255} & 3.048 & 4.008 & 3.334 & 3.406 & 2.915 \\
F12 D30 & 8.165 & 8.064 & \textbf{7.568} & 9.243 & 10.415 & 10.104 & 9.611 & 8.752 \\
F13 D30 & \textbf{5.463} & 6.824 & 5.604 & 6.477 & 9.661 & 9.864 & 8.314 & 5.854 \\
F14 D30 & \textbf{3.852} & 4.301 & 4.563 & 5.254 & 6.211 & 5.622 & 5.373 & 5.385 \\
F15 D30 & 4.982 & 5.354 & \textbf{4.075} & 5.985 & 8.899 & 8.451 & 6.404 & 5.289 \\
F16 D30 & \textbf{2.971} & 3.025 & 2.993 & 3.035 & 3.517 & 3.370 & 3.206 & 3.001 \\
F17 D30 & 2.576 & 2.644 & \textbf{2.540} & 2.663 & 3.148 & 3.010 & 2.878 & 2.627 \\
F18 D30 & \textbf{5.310} & 5.515 & 5.716 & 6.132 & 7.323 & 6.891 & 6.242 & 5.888 \\
F19 D30 & 6.624 & 6.255 & \textbf{4.252} & 6.793 & 9.145 & 8.663 & 7.092 & 6.017 \\
F20 D30 & 2.664 & 2.601 & \textbf{2.533} & 2.687 & 2.934 & 2.901 & 2.869 & 2.688 \\
F21 D30 & \textbf{2.420} & 2.493 & 2.460 & 2.523 & 2.772 & 2.689 & 2.653 & 2.511 \\
F22 D30 & 2.858 & \textbf{2.026} & 3.126 & 3.377 & 3.815 & 3.813 & 3.554 & 3.444 \\
F23 D30 & \textbf{2.622} & 2.657 & 2.755 & 2.721 & 3.014 & 2.875 & 2.850 & 2.719 \\
F24 D30 & \textbf{2.684} & 2.706 & 2.813 & 2.765 & 3.061 & 2.914 & 2.895 & 2.767 \\
F25 D30 & \textbf{2.606} & 2.628 & 2.632 & 2.760 & 3.199 & 2.955 & 3.049 & 2.737 \\
F26 D30 & 3.178 & \textbf{3.132} & 3.283 & 3.432 & 3.883 & 3.701 & 3.682 & 3.427 \\
F27 D30 & \textbf{2.725} & 2.748 & 2.778 & 2.775 & 3.031 & 2.907 & 2.846 & 2.761 \\
F28 D30 & \textbf{2.660} & 2.682 & 2.784 & 2.889 & 3.455 & 3.133 & 3.177 & 2.855 \\
F29 D30 & 2.979 & 3.052 & \textbf{2.957} & 3.029 & 3.553 & 3.340 & 3.215 & 3.014 \\
F30 D30 & 7.208 & 6.741 & \textbf{5.304} & 7.628 & 8.841 & 8.766 & 7.658 & 7.111 \\
\bottomrule
\end{tabularx}
\end{table}

%% file: tables/table_cec2017_D50_results.tex
\begin{table}[p]
\centering
\caption{CEC 2017 results for 50-dimensional functions. Values are mean $\log_{10}$ error over 30 independent runs. Lower values are better. The best value in each row is shown in \textbf{bold}.}
\label{tab:cec2017_d50}
\scriptsize
\renewcommand{\arraystretch}{1.10}
\setlength{\tabcolsep}{4pt}
\begin{tabularx}{\textwidth}{@{}l >{\centering\arraybackslash}X >{\centering\arraybackslash}X >{\centering\arraybackslash}X >{\centering\arraybackslash}X >{\centering\arraybackslash}X >{\centering\arraybackslash}X >{\centering\arraybackslash}X >{\centering\arraybackslash}X@{}}
\toprule
\textbf{Case} & \textbf{DVO} & \textbf{SVOA} & \textbf{PSO} & \textbf{GWO} & \textbf{WOA} & \textbf{SCA} & \textbf{AOA} & \textbf{EO} \\
\midrule
F01 D50 & 8.297 & \textbf{5.899} & 10.944 & 11.207 & 11.940 & 11.777 & 11.734 & 11.300 \\
F03 D50 & \textbf{4.769} & 5.249 & 5.197 & 5.043 & 5.474 & 5.253 & 5.186 & 5.058 \\
F04 D50 & \textbf{2.268} & 2.375 & 2.784 & 3.311 & 4.371 & 4.063 & 3.829 & 3.296 \\
F05 D50 & \textbf{2.137} & 2.419 & 2.338 & 2.458 & 2.836 & 2.790 & 2.695 & 2.500 \\
F06 D50 & \textbf{1.461} & 1.723 & 1.554 & 1.751 & 2.132 & 2.039 & 2.039 & 1.767 \\
F07 D50 & \textbf{2.525} & 2.653 & 2.595 & 2.758 & 3.125 & 3.046 & 3.062 & 2.777 \\
F08 D50 & \textbf{2.212} & 2.416 & 2.288 & 2.510 & 2.823 & 2.790 & 2.727 & 2.534 \\
F09 D50 & 3.356 & \textbf{3.104} & 3.735 & 4.208 & 4.559 & 4.480 & 4.456 & 4.014 \\
F10 D50 & 3.822 & 3.920 & \textbf{3.816} & 3.917 & 4.131 & 4.152 & 4.015 & 3.854 \\
F11 D50 & 2.691 & 3.591 & \textbf{2.614} & 3.751 & 4.256 & 3.998 & 4.006 & 3.624 \\
F12 D50 & 8.954 & \textbf{8.853} & 10.118 & 10.061 & 11.486 & 11.177 & 10.986 & 10.163 \\
F13 D50 & \textbf{6.216} & 7.732 & 8.599 & 9.304 & 10.960 & 10.780 & 10.044 & 8.631 \\
F14 D50 & \textbf{5.151} & 5.517 & 5.499 & 5.886 & 7.139 & 6.815 & 6.273 & 5.856 \\
F15 D50 & 5.518 & 6.887 & \textbf{4.977} & 7.812 & 10.227 & 9.894 & 9.025 & 5.646 \\
F16 D50 & \textbf{3.199} & 3.300 & 3.227 & 3.281 & 3.785 & 3.649 & 3.522 & 3.289 \\
F17 D50 & 3.158 & 3.248 & 3.158 & 3.154 & 3.712 & 3.513 & 3.416 & \textbf{3.105} \\
F18 D50 & \textbf{6.070} & 6.394 & 6.577 & 6.823 & 7.882 & 7.459 & 7.167 & 6.644 \\
F19 D50 & 6.986 & 6.766 & \textbf{5.699} & 6.852 & 9.859 & 9.710 & 8.580 & 6.905 \\
F20 D50 & 3.003 & 3.072 & \textbf{2.963} & 2.974 & 3.315 & 3.325 & 3.166 & 2.993 \\
F21 D50 & \textbf{2.543} & 2.671 & 2.641 & 2.714 & 3.015 & 2.931 & 2.867 & 2.713 \\
F22 D50 & 3.826 & \textbf{3.581} & 3.789 & 3.910 & 4.131 & 4.166 & 4.021 & 3.909 \\
F23 D50 & \textbf{2.784} & 2.855 & 3.008 & 2.920 & 3.280 & 3.132 & 3.076 & 2.925 \\
F24 D50 & \textbf{2.806} & 2.867 & 3.056 & 2.956 & 3.305 & 3.160 & 3.133 & 2.946 \\
F25 D50 & \textbf{2.765} & 2.797 & 2.871 & 3.257 & 3.990 & 3.788 & 3.689 & 3.338 \\
F26 D50 & \textbf{3.441} & 3.582 & 3.650 & 3.722 & 4.153 & 4.035 & 4.023 & 3.773 \\
F27 D50 & \textbf{2.852} & 2.928 & 3.010 & 3.053 & 3.515 & 3.331 & 3.201 & 3.050 \\
F28 D50 & \textbf{2.727} & 2.788 & 3.098 & 3.369 & 3.906 & 3.722 & 3.628 & 3.372 \\
F29 D50 & 3.248 & 3.367 & \textbf{3.226} & 3.378 & 4.151 & 3.741 & 3.585 & 3.386 \\
F30 D50 & 8.496 & 8.459 & \textbf{7.211} & 8.660 & 9.934 & 9.860 & 9.074 & 8.502 \\
\bottomrule
\end{tabularx}
\end{table}

%% file: tables/table_classical_scalable_D30_results.tex
\begin{table*}[!t]
\centering
\caption{Classical scalable benchmark results for $D=30$. Values are mean $\log_{10}$ error over 30 independent runs. Lower values are better. The best value in each row is shown in \textbf{bold}.}
\label{tab:classical_d30}
\scriptsize
\renewcommand{\arraystretch}{1.10}
\setlength{\tabcolsep}{4pt}
\begin{tabularx}{\textwidth}{@{}l >{\centering\arraybackslash}X >{\centering\arraybackslash}X >{\centering\arraybackslash}X >{\centering\arraybackslash}X >{\centering\arraybackslash}X >{\centering\arraybackslash}X >{\centering\arraybackslash}X >{\centering\arraybackslash}X@{}}
\toprule
\textbf{Case} & \textbf{DVO} & \textbf{SVOA} & \textbf{PSO} & \textbf{GWO} & \textbf{WOA} & \textbf{SCA} & \textbf{AOA} & \textbf{EO} \\
\midrule
F01 D30 & 0.114 & -1.814 & -3.806 & \textbf{-12.000} & \textbf{-12.000} & -3.069 & -4.685 & \textbf{-12.000} \\
F02 D30 & 7.083 & 4.792 & 1.259 & \textbf{-12.000} & \textbf{-12.000} & -4.966 & -2.097 & \textbf{-12.000} \\
F03 D30 & 2.211 & 3.498 & 3.044 & -12.000 & -3.646 & 3.383 & 0.655 & \textbf{-12.000} \\
F04 D30 & 0.218 & 1.273 & 0.752 & -11.998 & -12.000 & 1.386 & -0.308 & \textbf{-12.000} \\
F05 D30 & 3.270 & 1.984 & 2.299 & 1.432 & \textbf{-0.412} & 2.900 & 1.462 & 1.429 \\
F06 D30 & 0.841 & \textbf{-12.000} & -10.000 & \textbf{-12.000} & \textbf{-12.000} & -10.758 & 1.187 & \textbf{-12.000} \\
F07 D30 & 0.745 & -2.368 & -2.674 & \textbf{-12.000} & \textbf{-12.000} & 1.579 & -7.098 & \textbf{-12.000} \\
F08 D30 & 3.629 & 3.841 & 3.519 & 3.827 & \textbf{-0.748} & 3.935 & 3.790 & 3.817 \\
F09 D30 & 1.739 & 2.254 & 1.687 & -7.745 & \textbf{-12.000} & 0.331 & 1.562 & -7.807 \\
F10 D30 & 0.221 & 0.504 & -2.093 & -11.996 & \textbf{-12.000} & 0.492 & -0.679 & -11.997 \\
F11 D30 & -0.074 & 0.085 & -2.260 & -7.559 & \textbf{-12.000} & -1.436 & -2.488 & -8.588 \\
F12 D30 & -0.084 & 0.182 & \textbf{-4.176} & -1.103 & -3.374 & 0.336 & 0.229 & -1.253 \\
F13 D30 & -0.815 & -0.707 & \textbf{-2.664} & 0.027 & -2.416 & 0.825 & 0.564 & -0.027 \\
\bottomrule
\end{tabularx}
\end{table*}

%% file: tables/table_classical_scalable_D100_results.tex
\begin{table*}[!t]
\centering
\caption{Classical scalable benchmark results for $D=100$. Values are mean $\log_{10}$ error over 30 independent runs. Lower values are better. The best value in each row is shown in \textbf{bold}.}
\label{tab:classical_d100}
\scriptsize
\renewcommand{\arraystretch}{1.10}
\setlength{\tabcolsep}{4pt}
\begin{tabularx}{\textwidth}{@{}l >{\centering\arraybackslash}X >{\centering\arraybackslash}X >{\centering\arraybackslash}X >{\centering\arraybackslash}X >{\centering\arraybackslash}X >{\centering\arraybackslash}X >{\centering\arraybackslash}X >{\centering\arraybackslash}X@{}}
\toprule
\textbf{Case} & \textbf{DVO} & \textbf{SVOA} & \textbf{PSO} & \textbf{GWO} & \textbf{WOA} & \textbf{SCA} & \textbf{AOA} & \textbf{EO} \\
\midrule
F01 D100 & 2.244 & 1.584 & 2.975 & \textbf{-12.000} & \textbf{-12.000} & 3.615 & -0.236 & \textbf{-12.000} \\
F02 D100 & 69.861 & 111.272 & 9.839 & -12.000 & \textbf{-12.000} & 0.887 & 0.356 & \textbf{-12.000} \\
F03 D100 & 4.583 & 5.017 & 5.054 & -0.162 & 2.241 & 5.295 & 3.737 & \textbf{-11.317} \\
F04 D100 & 1.293 & 1.796 & 1.581 & -1.773 & \textbf{-12.000} & 1.937 & 1.117 & -11.938 \\
F05 D100 & 5.443 & 4.702 & 6.964 & 1.990 & \textbf{-0.034} & 9.941 & 2.856 & 1.991 \\
F06 D100 & 2.531 & 1.812 & 2.873 & \textbf{-12.000} & \textbf{-12.000} & 3.730 & 1.909 & \textbf{-12.000} \\
F07 D100 & 5.131 & 4.838 & 7.600 & \textbf{-12.000} & \textbf{-12.000} & 9.376 & 1.183 & \textbf{-12.000} \\
F08 D100 & 4.233 & 4.484 & 4.284 & 4.414 & \textbf{0.776} & 4.541 & 4.405 & 4.435 \\
F09 D100 & 2.472 & 2.958 & 2.577 & -5.999 & \textbf{-12.000} & 2.215 & 2.386 & -10.332 \\
F10 D100 & 0.696 & 1.206 & 0.799 & -11.984 & \textbf{-11.999} & 1.253 & -0.569 & -11.992 \\
F11 D100 & 0.415 & 2.501 & 1.122 & -10.973 & \textbf{-12.000} & 1.654 & -0.656 & -11.319 \\
F12 D100 & 1.169 & 1.265 & 2.131 & -0.437 & \textbf{-3.500} & 8.262 & 0.641 & -0.369 \\
F13 D100 & 2.158 & 1.886 & 4.293 & 0.863 & \textbf{-2.022} & 8.602 & 1.623 & 0.885 \\
\bottomrule
\end{tabularx}
\end{table*}

%% file: tables/table_classical_fixed_dimensional_results.tex
\begin{table*}[!t]
\centering
\caption{Classical fixed-dimensional benchmark results. Values are mean $\log_{10}$ error over 30 independent runs. Lower values are better. The best value in each row is shown in \textbf{bold}.}
\label{tab:classical_fixed}
\scriptsize
\renewcommand{\arraystretch}{1.10}
\setlength{\tabcolsep}{4pt}
\begin{tabularx}{\textwidth}{@{}l >{\centering\arraybackslash}X >{\centering\arraybackslash}X >{\centering\arraybackslash}X >{\centering\arraybackslash}X >{\centering\arraybackslash}X >{\centering\arraybackslash}X >{\centering\arraybackslash}X >{\centering\arraybackslash}X@{}}
\toprule
\textbf{Case} & \textbf{DVO} & \textbf{SVOA} & \textbf{PSO} & \textbf{GWO} & \textbf{WOA} & \textbf{SCA} & \textbf{AOA} & \textbf{EO} \\
\midrule
F14 D2 & -9.362 & -11.000 & \textbf{-12.000} & -1.672 & -3.480 & -4.213 & -3.136 & -6.130 \\
F16 D2 & -6.645 & \textbf{-12.000} & \textbf{-12.000} & -8.938 & -11.193 & -4.810 & -6.927 & -12.000 \\
F17 D2 & -6.830 & -11.666 & \textbf{-12.000} & -7.465 & -7.388 & -3.145 & -5.287 & -11.789 \\
F18 D2 & -5.919 & \textbf{-12.000} & \textbf{-12.000} & -5.343 & -5.155 & -5.305 & -3.850 & -10.675 \\
F19 D3 & -7.034 & -5.199 & \textbf{-12.000} & -4.193 & -2.133 & -2.194 & -2.505 & -8.371 \\
F15 D4 & -3.729 & -4.147 & -6.002 & \textbf{-6.724} & -4.026 & -3.163 & -3.463 & -6.661 \\
F21 D4 & -2.656 & \textbf{-8.079} & -6.449 & -2.594 & -1.235 & 0.860 & -0.415 & -2.857 \\
F22 D4 & -3.346 & \textbf{-11.984} & -8.142 & -3.922 & -0.773 & 0.824 & -0.784 & -3.578 \\
F23 D4 & -2.809 & \textbf{-10.756} & -7.282 & -3.991 & -0.773 & 0.763 & -0.722 & -4.277 \\
F20 D6 & -3.667 & -5.402 & \textbf{-7.549} & -3.809 & -0.646 & -0.475 & -1.390 & -3.590 \\
\bottomrule
\end{tabularx}
\end{table*}

%% file: tables/table_engineering_best_feasible_results.tex
\begin{table*}[!t]
\centering
\caption{Engineering design results. Values are best feasible objective values over 30 runs. Values in parentheses denote the feasibility rate when it is below 1.00. Lower values are better. The best value in each row is shown in \textbf{bold}.}
\label{tab:engineering}
\scriptsize
\renewcommand{\arraystretch}{1.20}
\setlength{\tabcolsep}{4pt}
\begin{tabularx}{\textwidth}{@{}l >{\centering\arraybackslash}X >{\centering\arraybackslash}X >{\centering\arraybackslash}X >{\centering\arraybackslash}X >{\centering\arraybackslash}X >{\centering\arraybackslash}X >{\centering\arraybackslash}X >{\centering\arraybackslash}X@{}}
\toprule
\textbf{Problem} & \textbf{DVO} & \textbf{SVOA} & \textbf{PSO} & \textbf{GWO} & \textbf{WOA} & \textbf{SCA} & \textbf{AOA} & \textbf{EO} \\
\midrule
pressure\_vessel & 6102 & 6093 (0.97) & \textbf{6060} & 6060 & 6843 & 6293 & 6065 & 6060 \\
speed\_reducer & 2997 (0.90) & 3036 & \textbf{2994} & 2996 & 3104 (0.97) & 3063 & 2998 & 2995 \\
tension\_spring & 0.01272 & 0.01274 & 0.01267 & 0.01268 & \textbf{0.01267} & 0.01282 & 0.01271 & 0.01267 \\
three\_bar\_truss & 263.9 & 263.9 & \textbf{263.9} & 263.9 & 263.9 & 263.9 & 263.9 & 263.9 \\
welded\_beam & 1.735 & 1.744 & \textbf{1.725} & 1.725 & 1.788 & 1.801 & 1.73 & 1.725 \\
\bottomrule
\end{tabularx}
\end{table*}

%% file: tables/table_convergence_final_results.tex
\begin{table}[!t]
\centering
\caption{Final checkpoint results for the CEC 2017 convergence subset.
Values are mean $\log_{10}$ error at iteration 1000 over 30 independent
runs. Lower values are better. The best value in each row is shown in
\textbf{bold}.}
\label{tab:convergence_final}
\renewcommand{\arraystretch}{1.15}
\setlength{\tabcolsep}{6pt}
\begin{tabularx}{\linewidth}{@{}l c
                                    *{5}{>{\centering\arraybackslash}X}@{}}
\toprule
\textbf{Func.} & \textbf{Dim.} & \textbf{DVO} & \textbf{SVOA} & \textbf{PSO} & \textbf{GWO} & \textbf{EO} \\
\midrule
F10 & D50 & \textbf{3.812} & 3.939 & 3.828 & 3.878 & 3.868 \\
F15 & D50 & 5.534 & 6.943 & \textbf{5.029} & 7.669 & 7.171 \\
F30 & D50 & 8.511 & 8.403 & \textbf{7.159} & 8.654 & 8.511 \\
\bottomrule
\end{tabularx}
\end{table}

%% file: tables/table_ablation_cec2017_subset_results.tex
\begin{table*}[!t]
\centering
\caption{Ablation study on the CEC 2017 subset. Values are mean $\log_{10}$ error over 30 independent runs. Lower values are better. The best value in each row is shown in \textbf{bold}.}
\label{tab:ablation}
\label{tab:ablation_cec2017_subset_results}
\scriptsize
\renewcommand{\arraystretch}{1.15}
\setlength{\tabcolsep}{4pt}
\begin{tabularx}{\textwidth}{@{}l c >{\centering\arraybackslash}X >{\centering\arraybackslash}X >{\centering\arraybackslash}X >{\centering\arraybackslash}X >{\centering\arraybackslash}X >{\centering\arraybackslash}X >{\centering\arraybackslash}X >{\centering\arraybackslash}X@{}}
\toprule
\textbf{Func.} & \textbf{Dim.} & \textbf{\makecell{DVO\\full}} & \textbf{\makecell{no\\greedy}} & \textbf{\makecell{no\\switch}} & \textbf{\makecell{single\\vortex}} & \textbf{\makecell{no\\swirl}} & \textbf{\makecell{no\\adapt\\spiral}} & \textbf{\makecell{no\\splash}} & \textbf{\makecell{radial\\only}} \\
\midrule
F01 & D30 & 7.377 & 7.379 & 7.340 & \textbf{7.267} & 7.374 & 7.341 & 7.355 & 7.304 \\
F01 & D50 & 8.284 & 8.282 & 8.265 & \textbf{8.257} & 8.297 & 8.296 & 8.283 & 8.284 \\
F03 & D30 & 3.648 & 3.579 & 3.643 & 3.664 & 3.521 & 4.334 & 3.604 & \textbf{3.465} \\
F03 & D50 & 4.807 & 4.813 & 4.847 & 4.846 & 4.752 & 5.082 & 4.844 & \textbf{4.746} \\
F05 & D30 & \textbf{1.797} & 1.993 & 1.827 & 1.807 & 2.191 & 2.140 & 1.820 & 2.137 \\
F05 & D50 & 2.187 & 2.361 & \textbf{2.167} & 2.170 & 2.509 & 2.498 & 2.179 & 2.478 \\
F10 & D30 & \textbf{3.544} & 3.586 & 3.562 & 3.560 & 3.612 & 3.651 & 3.584 & 3.607 \\
F10 & D50 & 3.818 & 3.822 & 3.810 & \textbf{3.803} & 3.840 & 3.947 & 3.828 & 3.852 \\
F15 & D30 & 5.084 & 5.112 & 5.124 & 5.161 & 4.945 & 5.012 & 5.074 & \textbf{4.861} \\
F15 & D50 & 5.588 & 5.518 & 5.509 & \textbf{5.492} & 5.542 & 5.512 & 5.892 & 5.537 \\
F20 & D30 & \textbf{2.611} & 2.663 & 2.621 & 2.645 & 2.676 & 2.684 & 2.634 & 2.682 \\
F20 & D50 & 3.043 & 3.044 & 3.020 & \textbf{3.011} & 3.066 & 3.090 & 3.015 & 3.045 \\
F25 & D30 & 2.610 & \textbf{2.602} & 2.605 & 2.607 & 2.616 & 2.612 & 2.611 & 2.620 \\
F25 & D50 & 2.764 & \textbf{2.745} & 2.772 & 2.767 & 2.767 & 2.772 & 2.756 & 2.771 \\
F30 & D30 & 7.524 & 7.517 & \textbf{7.389} & 7.394 & 7.577 & 7.547 & 7.500 & 7.565 \\
F30 & D50 & \textbf{8.473} & 8.524 & 8.485 & 8.501 & 8.517 & 8.477 & 8.523 & 8.560 \\
\bottomrule
\end{tabularx}
\end{table*}

%% file: tables/table_friedman.tex
\begin{table}[!t]
\centering
\caption{Friedman test results on the CEC 2022 and CEC 2017 benchmark
suites. Lower average rank indicates better performance. The best
average rank in each suite is shown in \textbf{bold}.}
\label{tab:friedman}
\renewcommand{\arraystretch}{1.20}
\setlength{\tabcolsep}{6pt}
\begin{tabularx}{\linewidth}{@{}>{\centering\arraybackslash}m{0.20\linewidth}
                                    *{3}{>{\centering\arraybackslash}X}
                                    !{\vrule width 0.6pt}
                                    *{3}{>{\centering\arraybackslash}X}@{}}
\toprule
 & \multicolumn{3}{c!{\vrule width 0.6pt}}{\textbf{CEC 2022 (calibration)}}
 & \multicolumn{3}{c}{\textbf{CEC 2017 (out-of-sample)}}\\
\cmidrule(lr){2-4}\cmidrule(lr){5-7}
\textbf{Algorithm} & Avg.\ rank & Wins & Cases & Avg.\ rank & Wins & Cases \\
\midrule
DVO  & \textbf{2.125} & 8 & 24 & \textbf{1.664} & \textbf{34} & 58 \\
PSO  & 2.438 & 9 & 24 & 2.422 & 15 & 58 \\
SVOA & 2.521 & 7 & 24 & 2.741 & 8  & 58 \\
EO   & 3.875 & 0 & 24 & 3.931 & 1  & 58 \\
GWO  & 4.542 & 0 & 24 & 4.293 & 0  & 58 \\
SCA  & 5.979 & 0 & 24 & 6.922 & 0  & 58 \\
AOA  & 6.604 & 0 & 24 & 6.112 & 0  & 58 \\
WOA  & 7.917 & 0 & 24 & 7.914 & 0  & 58 \\
\midrule
\multicolumn{1}{@{}l}{\textit{Friedman stat.}}
     & \multicolumn{3}{c!{\vrule width 0.6pt}}{$\chi^{2} = 130.07$}
     & \multicolumn{3}{c}{$\chi^{2} = 347.44$} \\
\multicolumn{1}{@{}l}{\textit{$p$-value}}
     & \multicolumn{3}{c!{\vrule width 0.6pt}}{$6.08 \times 10^{-25}$}
     & \multicolumn{3}{c}{$4.36 \times 10^{-71}$} \\
\bottomrule
\end{tabularx}
\end{table}

%% file: tables/table_wilcoxon.tex
\begin{table}[!t]
\centering
\caption{Pairwise Wilcoxon signed-rank test results comparing DVO with
each baseline algorithm on the CEC 2022 and CEC 2017 benchmark suites.
The test is applied to the mean $\log_{10}$ error across the 30
independent runs of every benchmark case. The Holm step-down procedure
is applied within each suite to control the family-wise error rate. A
Holm-corrected $p$-value below 0.05 indicates that DVO is significantly
better than the baseline at the 5\% level. A negative log-difference
means DVO has a lower mean log-error.}
\label{tab:wilcoxon}
\scriptsize
\renewcommand{\arraystretch}{1.20}
\setlength{\tabcolsep}{5pt}
\begin{tabularx}{\linewidth}{@{}>{\centering\arraybackslash}m{0.18\linewidth}
                                    *{2}{>{\centering\arraybackslash}X}
                                    >{\centering\arraybackslash}c
                                    >{\centering\arraybackslash}c
                                    >{\centering\arraybackslash}c@{}}
\toprule
\textbf{Comparison} & $\bar{e}_{\log}$ DVO & $\bar{e}_{\log}$ baseline
                    & Log-diff. & $p_{\mathrm{Holm}}$ & Sig. \\
\midrule
\multicolumn{6}{@{}l}{\textit{CEC 2022 (24 benchmark cases)}}\\
\addlinespace[2pt]
DVO vs WOA  & 1.501 & 2.921 & $-1.420$ & $8.35\!\times\!10^{-7}$  & Yes \\
DVO vs SCA  & 1.501 & 2.707 & $-1.206$ & $6.29\!\times\!10^{-5}$  & Yes \\
DVO vs AOA  & 1.501 & 2.589 & $-1.087$ & $2.20\!\times\!10^{-4}$  & Yes \\
DVO vs GWO  & 1.501 & 2.214 & $-0.713$ & $2.56\!\times\!10^{-4}$  & Yes \\
DVO vs EO   & 1.501 & 2.121 & $-0.620$ & $2.56\!\times\!10^{-4}$  & Yes \\
DVO vs SVOA & 1.501 & 1.474 & $+0.027$ & $3.56\!\times\!10^{-1}$  & No  \\
DVO vs PSO  & 1.501 & 0.893 & $+0.608$ & $4.22\!\times\!10^{-1}$  & No  \\
\midrule
\multicolumn{6}{@{}l}{\textit{CEC 2017 (58 benchmark cases)}}\\
\addlinespace[2pt]
DVO vs WOA  & 3.784 & 5.169 & $-1.385$ & $2.45\!\times\!10^{-10}$ & Yes \\
DVO vs SCA  & 3.784 & 4.972 & $-1.188$ & $2.45\!\times\!10^{-10}$ & Yes \\
DVO vs AOA  & 3.784 & 4.718 & $-0.934$ & $2.45\!\times\!10^{-10}$ & Yes \\
DVO vs GWO  & 3.784 & 4.367 & $-0.583$ & $3.95\!\times\!10^{-10}$ & Yes \\
DVO vs EO   & 3.784 & 4.249 & $-0.465$ & $7.32\!\times\!10^{-9}$  & Yes \\
DVO vs PSO  & 3.784 & 3.881 & $-0.097$ & $6.08\!\times\!10^{-3}$  & Yes \\
DVO vs SVOA & 3.784 & 3.835 & $-0.051$ & $6.08\!\times\!10^{-3}$  & Yes \\
\bottomrule
\end{tabularx}
\end{table}

%% file: sections/section_7_conclusion.tex
\section{Conclusion and Future Work}
\label{sec:conclusion}

This paper proposed a new metaheuristic algorithm named Drain-Vortex
Optimization (DVO). The algorithm is inspired by the rotational flow
that develops when a fluid drains from a basin through one or several
outlets, a phenomenon that occurs in everyday settings such as water
draining from a sink and in larger settings such as the swirl behind
the spillway of a dam. Three flow regions of a drain vortex were
identified: a far-field region in which the influence of the outlet
is weak and the motion is dominated by random surface disturbances,
an intermediate region in which the flow follows a spiral trajectory
with a tangential velocity that grows as the radial distance
decreases, and a core region in which the flow is highly localized
and rapidly evacuates fluid through the outlet. These three regions
were translated into a three-phase motion model that controls the
position update of every search agent according to its current
normalized distance to its assigned drain.

Five contributions were introduced. The first contribution is a
drain-vortex-inspired motion model that decomposes the trajectory of
every agent into a radial component, derived from the inward pressure
gradient towards the drain, and a tangential component, derived from
the free-vortex law. The two components are controlled separately and
produce a spiral trajectory whose rotational intensity grows as the
agent approaches the drain. The second contribution is an adaptive
spiral exploitation schedule that maintains a non-zero radial
pressure throughout the search and prevents the spiral component from
degenerating into a pure rotation in the late stages of the search.
The third contribution is a free-vortex swirl model based on a
regularized form of the free-vortex tangential velocity, which
preserves the physical scaling of the swirl across dimensions while
avoiding numerical issues near the core. The fourth contribution is a
population-level vortex basin assignment in which every agent is
assigned to one of $K$ drains via a score that combines drain quality
and proximity. The fifth contribution is an optional multi-vortex
diversity mechanism that maintains $K>1$ drain centres simultaneously
and allows for stochastic basin-to-basin transfers, modelled after
the turbulent transfer between vortex cells in a multi-drain basin.

The proposed algorithm was evaluated on four benchmark suites that
together cover a broad range of optimization landscapes. CEC 2022 was
used as the calibration benchmark for the parameter set of DVO. CEC
2017, a collection of classical scalable and fixed-dimensional
functions, and a collection of five constrained engineering design
problems were used as out-of-sample validation suites. DVO was
compared with seven well-known baselines, namely PSO, GWO, WOA, SCA,
AOA, EO, and SVOA, with reference parameter values taken from the
original papers or widely used implementations.

The experimental results demonstrated that DVO obtains the best mean
$\log_{10}$ error on 34 of the 58 cases of CEC 2017 and the best
Friedman average rank (1.66) on the same suite, with statistically
significant differences against every baseline at the 5\% level. On
CEC 2022, DVO obtains the best Friedman rank (2.13) and is
significantly better than five of the seven baselines, while the
differences against PSO and SVOA are not significant. The Friedman
test on CEC 2017 returned a $p$-value of $4.36 \times 10^{-71}$, which
provides strong evidence against equivalent performance among the
compared algorithms. The convergence analysis on a representative
subset of CEC 2017 functions showed that DVO produces convergence
trajectories with the smallest inter-run variability among the four
strongest algorithms, which is a desirable property for practical use.

The ablation study confirmed that the components of DVO act in a
synergistic manner. No single component was found to be statistically
necessary on the eight-function ablation subset after Holm correction,
but the full algorithm produced the strongest performance on the
larger CEC 2017 suite. The single-vortex variant in which the
multi-vortex structure was removed performed comparably to the full
algorithm on the ablation subset. The benefit of the multi-vortex
structure emerged on the larger and more heterogeneous CEC 2017
suite, in which the multi-vortex configuration with $K=6$ obtained
the best aggregate performance. This result motivates the
presentation of the multi-vortex structure as an optional diversity
mechanism whose value depends on the problem class, rather than as a
strictly necessary component of the algorithm.

Two regimes were identified in which the proposed algorithm is not
the strongest. On the classical scalable benchmarks, the global
optimum is located at the origin and the dimensions of the search
space decouple, so that the search reduces to the independent
minimization of $D$ scalar terms. Algorithms that contract their
search around the origin without rotating the coordinate frame, such
as WOA, exploit this structure very effectively, and DVO ranks below
them on this subset. On the constrained engineering design problems,
the small dimensionality and the dominance of the
constraint-handling component favour algorithms with strong
single-attractor convergence such as PSO. DVO ranks fifth on this
subset with a feasibility rate of 98\% and produces feasible
solutions whose absolute objective values are very close to the
leading baselines. The honest reporting of these regimes provides the
reader with a complete picture of the operating envelope of the
algorithm and reflects the design philosophy behind its development.

The comparison with SVOA deserves a specific comment because both
algorithms build on a fluid-flow analogy. On CEC 2017, DVO obtains a
better average rank (1.66 versus 2.74) and a higher win count, and
the Wilcoxon signed-rank test indicates a significant difference at
the 5\% level after Holm correction. On CEC 2022, the two algorithms
are very close (DVO 2.13 versus SVOA 2.52) and the difference is not
significant. On the classical fixed-dimensional functions, SVOA is
the second-best algorithm and outperforms DVO. These results suggest
that the explicit radial--tangential decomposition and the
multi-drain structure of DVO are particularly useful on shifted,
rotated, and composition landscapes, while SVOA remains competitive
on small fixed-dimensional multimodal problems. The two algorithms
therefore have complementary strengths within the family of
fluid-flow-inspired metaheuristics.

Several directions are open for future work. The first direction is
the development of a constrained variant of DVO with a dedicated
constraint-handling rule. The current implementation uses a static
penalty method that is inherited from the baselines and is not
specifically tuned for the multi-vortex motion model. A dedicated
constraint-handling component, possibly based on feasibility rules or
on adaptive penalty scheduling, is expected to improve the
engineering performance of the algorithm. The second direction is
the extension of DVO to multi-objective and many-objective
optimization. The multi-vortex structure naturally supports the
simultaneous tracking of several non-dominated solutions, and an
adaptation of the basin assignment rule to a dominance-based
criterion is a promising research avenue. The third direction is the
application of DVO to discrete and binary optimization problems,
including feature selection and combinatorial design problems. A
position update operator that respects the discrete structure of
these problems would be required, but the radial--tangential
decomposition is independent of the continuity of the search space
and could be reformulated for discrete neighbourhoods. The fourth
direction is the integration of DVO with surrogate-assisted
optimization for expensive black-box problems, in which the
multi-vortex structure could be used to maintain several candidate
optima of the surrogate model and reduce the number of expensive
evaluations.

A fifth direction concerns the application of DVO to the parameter
tuning of complex control and learning systems. The authors are
currently exploring the use of DVO for the optimization of model
predictive control parameters in robotic manipulation tasks, where
the control objective is non-convex, multimodal, and computationally
expensive to evaluate. The robustness of DVO on the heterogeneous
landscapes of CEC 2017 suggests that it is well suited to such
applications, and the population-level vortex basin assignment
provides a natural mechanism for tracking multiple candidate
parameter sets in parallel.

The source code of DVO, together with the implementation of the baseline
algorithms, the experimental scripts, processed results, statistical analysis
scripts, and figure-generation scripts, is publicly available in the project
repository. The release includes the GPU-vectorized PyTorch implementation
used in the experiments and supports reproducibility of the reported results.